\documentclass[10pt,twocolumn,letterpaper]{article}

\usepackage[accsupp]{axessibility} % Improves PDF readability for those with disabilities.
\usepackage[pagenumbers]{cvpr} % To force page numbers, e.g. for an arXiv version

\usepackage{array}
\usepackage{tabularx}
\usepackage{multirow}
\usepackage[dvipsnames]{xcolor}
\usepackage[inkscapelatex=false]{svg}

\definecolor{cvprblue}{rgb}{0.21,0.49,0.74}
\usepackage[pagebackref,breaklinks,colorlinks,citecolor=cvprblue]{hyperref}

\def\paperID{15635} % *** Enter the Paper ID here
\def\confName{CVPR}
\def\confYear{2024}

\title{Open-Vocabulary Semantic Segmentation with Image Embedding Balancing}

\author{Xiangheng Shan, Dongyue Wu, Guilin Zhu, Yuanjie Shao$^{*}$, Nong Sang, Changxin Gao\\
	National Key Laboratory of Multispectral Information Intelligent Processing Technology,\\ School of Artificial Intelligence and Automation, Huazhong University of Science and Technology\\
 {\tt\small\{xianghengshan, dongyue\_wu, gzhu, shaoyuanjie, nsang, cgao\}@hust.edu.cn}
}

\begin{document}

\maketitle
\let\thefootnote\relax\footnotetext{$*$ Corresponding author.}

\begin{abstract}
Open-vocabulary semantic segmentation is a challenging task, which requires the model to output semantic masks of an image beyond a close-set vocabulary. 
Although many efforts have been made to utilize powerful CLIP models to accomplish this task, they are still easily overfitting to training classes due to the natural gaps in semantic information between training and new classes.
% Open-vocabulary semantic segmentation is a challenging task. Given an image, this task requires the model to output semantic masks corresponding to an arbitrary set of classes. 
%Many works attempt to leverage the powerful CLIP model for this task. 
% However, there are two challenges faced by CLIP-based models.
% However, they still face two challenges.
%However, they still face a challenge that training on semantic segmentation datasets often makes the model overfit to the classes in those datasets.
% Second, the low training image resolution and image-level supervision of CLIP lead to the lack of important spatial information of the image features from CLIP model.
% for this task 
%To overcome the first challenge, we design a decoder where we propose a xxx loss and a embedding fusion strategy. 
To overcome this challenge, we propose a novel framework for open-vocabulary semantic segmentation called EBSeg, incorporating an Adaptively Balanced Decoder (AdaB Decoder) and a Semantic Structure Consistency loss (SSC Loss). 
% The AdaB Decoder can produce different image embeddings for training and new classes and adaptively balance them to fully exploit their ability to model accurate semantic of training classes and excellent generalization ability for new classes.
The AdaB Decoder is designed to 
 generate different image embeddings for both training and new classes.
Subsequently, these two types of embeddings are adaptively balanced to fully exploit their ability to recognize training classes and generalization ability for new classes.
% The loss constrains distances between the image embeddings to be similar with the distances between the corresponding text embeddings, which allows our model to better learn the semantic structure of CLIP and improves the generalization ability of our model.
% To capture more semantic structure between classes, the SSC Loss align the inter-classes affinity in the image feature space with that in the text feature space of CLIP, allowing our model to learn a consistent semantic structure of CLIP and improves the generalization ability of our model.
To learn a consistent semantic structure from CLIP, the SSC Loss aligns the inter-classes affinity in the image feature space with that in the text feature space of CLIP, thereby improving the generalization ability of our model.
% Moreover, we produce different image embeddings for seen and unseen classes and adaptively fuse them to fully exploit their ability to modeling accurate semantic of seen classes and excellent generalization on unseen classes, respectively.
% Moreover, the decoder produces different types of image embeddings which are good at training or new classes, and we combine them to make our model perform better on both training and new classes. 
Furthermore, we employ a frozen SAM image encoder to complement the spatial information that CLIP features lack due to the low training image resolution and image-level supervision inherent in CLIP. Extensive experiments conducted across various benchmarks demonstrate that the proposed EBSeg outperforms the state-of-the-art methods.
% Our code and trained models will be publicly available.
Our code and trained models will be here: \href{https://github.com/slonetime/EBSeg}{https://github.com/slonetime/EBSeg}.
\end{abstract}
\section{Introduction}
\label{sec:intro}

Semantic segmentation is an important computer vision task that requires the model to identify a class label for each pixel in an image. For traditional (or fully supervised) semantic segmentation, models are trained and evaluated on a fixed dataset with a specific set of classes, and the test set has the same classes and image distribution as the training set. For this task, many effective methods \cite{long2015fully, noh2015learning, liang2015semantic, chen2018encoder, cheng2021per, kirillov2019panoptic, xiao2018unified, cheng2022masked, xie2021segformer} have been proposed, and they have significantly improved the prediction accuracy. However, models designed for this task always fail to segment images in the real world well, because they were only trained on specific datasets with fixed sets of classes.

\begin{figure}
    \centering
    % \includesvg[width=1\linewidth]{figures/overview.svg}
    \includegraphics[width=1\linewidth]{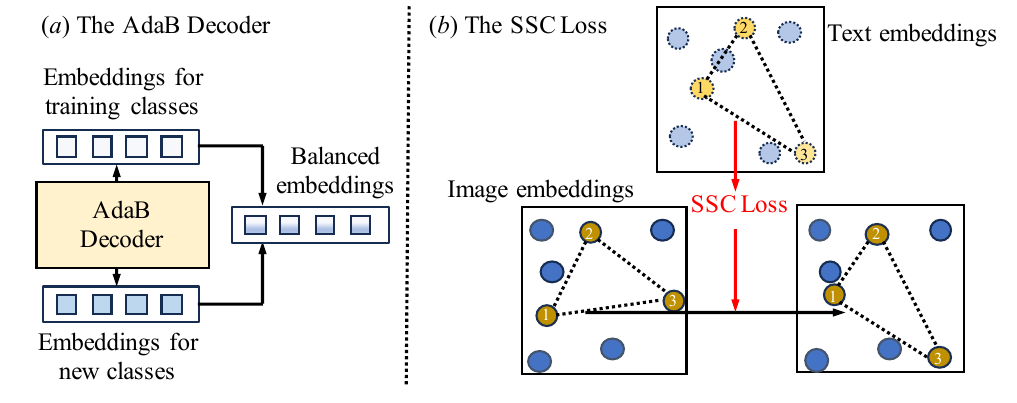}
    \caption{Illustration of our main idea. (\textit{a}) Our AdaB Decoder(Adaptively Balanced Decoder) outputs image embeddings for both training classes(classes existing in the training set during test) and new classes(classes not existing in training set). By adaptively balancing these embeddings, our model performs better at both training and new classes. (\textit{b}) We propose SSC Loss(Semantic Structure Consistency loss) that aligns the distribution of the image embeddings with that of the text embeddings. The SSC Loss helps our model learn the semantic structure of CLIP better and achieve better generalization capability for new classes.}
    \label{fig:overview}
\end{figure}

To address this problem, the open-vocabulary semantic segmentation task was introduced. In this task, given an image and an arbitrary set of classes, the model is expected to classify each pixel into its most corresponding class. This task is much closer to the real-world applications. Many works utilize the CLIP \cite{radford2021learning} model for this task because CLIP was trained on a large-scale image-text pair dataset and has strong generalization capability for open-vocabulary tasks.

Some existing works like \cite{li2022languagedriven, ghiasi2022scaling} propose to finetune models on semantic segmentation datasets. However, models finetuned on a semantic segmentation dataset often overfit to the training classes. 
Some others \cite{ding2021decoupling, xu2022simple, liang2023open} adopt a two-stage framework. Firstly, a category-agnostic mask generator is used to extract masks. Then, the masks are used to crop the input image to get many crops which will be fed into a frozen CLIP model for classification results. This framework incurs high computational costs as the CLIP model has to process many crops of the image. 
Since this framework does not leverage the discriminate features from CLIP in the mask generation process and lacks context in the classification process, it gets sub-optimal results.

Recently, some new methods adopting one-stage frameworks have emerged. ODISE \cite{xu2023open} employs a frozen diffusion model \cite{rombach2022high} to extract image features as the input for a Mask2former \cite{cheng2022masked} head. It \cite{xu2023open} leverages the strengths of both the diffusion model and the CLIP model to accomplish the open vocabulary segmentation task. However, ODISE faces high computational costs, as the diffusion model is very large (with about 1.5 billion parameters). MaskCLIP \cite{ding2023open} also uses a category-agnostic mask generator like \cite{ding2021decoupling, xu2022simple, liang2023open} but it \cite{ding2023open} does not crop the input image. 
MaskCLIP \cite{ding2023open} proposes a Relative Mask Attention module that applies the masks in the self-attention layers of CLIP image encoder as attention masks to produce mask attention embeddings for mask classification. 
SAN \cite{xu2023side} adds a lightweight image encoder to get masks and attention masks corresponding to each mask. Like MaskCLIP \cite{ding2023open}, the attention masks are fed to the last few layers of CLIP to obtain mask attention embeddings for mask classification. 

%Although these methods are effective, they still face two main challenges in this task.
%或多或少存在一些问题？不能很好的克服两个challenge？面对这个两个challenge？
% We summarize two main challenges faced by CLIP-based models for open-vocabulary semantic segmentation.
Although these methods are effective, they still face a challenge.
The challenge is that training on a specific semantic segmentation dataset often makes the model overfit to the training classes, impairing the generalization ability of the model, especially for large models.
To overcome the challenge, we present EBSeg (image Embedding Balancing for open-vocabulary semantic Segmentation). It consists of the Adaptively Balanced Decoder (AdaB Decoder) and the Semantic Structure Consistency loss (SSC Loss). 
% To overcome the first challenge, we propose the Adaptively Balanced Decoder (AdaB Decoder) and the Semantic Structure Consistency loss (SSC Loss). 
The AdaB Decoder generates mask attention embeddings, fully supervised embeddings specialized for training classes and frozen embeddings with excellent generalization for new classes. The mask attention embeddings come from a Mask Attention module introduced in MaskCLIP \cite{ding2023open} and SAN \cite{xu2023side}. The fully supervised embeddings are directly supervised by a cross-entropy loss with the training classes. The frozen embeddings are extracted from a frozen CLIP image encoder.
% XXXXX each one  xxxxx.
These three types of embeddings are then adaptively balanced to form a final representation of the input image for the final prediction.
Thus, AdaB Decoder could take full advantage of the superior features learned on training classes and the excellent generalizing ability on new classes at the same time.
On the other hand, SSC Loss aims at aligning the class-level affinity between the image features and text features.
Hence, our model could encode a more consistent class-level semantic structure from the CLIP feature space to enhance the generalization on new classes. 
% Additionally, we utilize a frozen SAM \cite{kirillov2023segment} image encoder to complement the class-agnostic spatial information of CLIP features to address the second issue that CLIP’s feature maps is bad for semantic segmentation.
Additionally, we utilize a frozen SAM \cite{kirillov2023segment} image encoder to complement the spatial information of CLIP features to address the issue that image feature maps of CLIP lack important spatial details for semantic segmentation.

We conduct extensive experiments on challenging open-vocabulary segmentation datasets to prove the effectiveness of our method EBSeg. Following previous works \cite{xu2022simple, xu2023side, liang2023open}, we train our model on COCO-Stuff \cite{caesar2018coco} and evaluate the model on VOC \cite{everingham2012pascal}, Pascal Context-59 \cite{mottaghi2014role}, Pascal Context-459 \cite{mottaghi2014role}, ADE20K-150 \cite{zhou2017scene} and ADE20K-847 \cite{zhou2017scene}. Our method achieves state-of-the-art results, with an average of 2.3\% mIoU improvements on these 5 benchmarks when using the CLIP ViT-B/16 model and an average of 2.3\% mIoU improvements with the CLIP ViT-L/14 model.

Our contributions are as follows:

\begin{itemize}
    % \item We summarize two main challenges faced by CLIP-based models for open-vocabulary semantic segmentation. That is, models training on semantic segmentation datasets often overfit to training classes and CLIP image features lacks the spatial information important for this task.
    %syj:你这个贡献我有点疑问，一个是过拟合的问题有人提过啊，另一个你说CLIP提取的视觉特征缺少空域信息，现在大部分的工作不就是针对这个问题才才用region-text pretraining来建立区域和文本的匹配关系。就是感觉你提出的两个挑战其实大家都有关注，不应该是你的贡献吧？
    %是的，我觉得作为贡献也不太好，那我就这一条去掉，然后把第二条拆成两条吧
    % \item For the first challenge faced by CLIP-based models, we propose Adaptively Balanced Decoder (AdaB Decoder). The AdaB Decoder adaptively balances different image embeddings to fully leverage their ability to recognize training classes and generalization capability for new classes.
    % \item We also propose Semantic Structure Consistency loss (SSC Loss) for the first challenge. The SSC Loss aligns the inter-classes affinity in the image feature space with that in the text feature space of CLIP, helping our model learn a consistent semantic structure from CLIP and improving the generalization ability of our model.
    % %The two methods help our model achieve better generalization capability for new classes.
    % \item To overcome the second challenge, we propose to leverage the powerful spatial information of features from SAM.
    % \item Our method EBSeg (image \textbf{E}mbedding \textbf{B}alancing open-vocabulary semantic \textbf{Seg}mentation) establishes a new state-of-the-art in the open-vocabulary semantic segmentation task.
    \item We propose Adaptively Balanced Decoder (AdaB Decoder). By adaptively balancing different image embeddings, AdaB Decoder can fully leverage their ability to recognize training classes and generalization capability for new classes that do not exist in the training set.
    \item We introduce the Semantic Structure Consistency loss (SSC Loss). The SSC Loss aligns the inter-classes affinity in the image feature space with that in the text feature space of CLIP. This loss helps our model learn a consistent semantic structure from CLIP and improves the generalization ability of our model.
    %The two methods help our model achieve better generalization capability for new classes.
    \item In our model, we propose to fuse the image features of SAM and CLIP to complement the spatial information of CLIP image features.
    \item Our method EBSeg establishes a new state-of-the-art in the open-vocabulary semantic segmentation task.
    
\end{itemize}

\section{Related works}
\label{sec:related}

\begin{figure*}
    \centering
    % \includesvg[width=\linewidth]{figures/model_architecture.svg}
     \includegraphics[width=\linewidth]{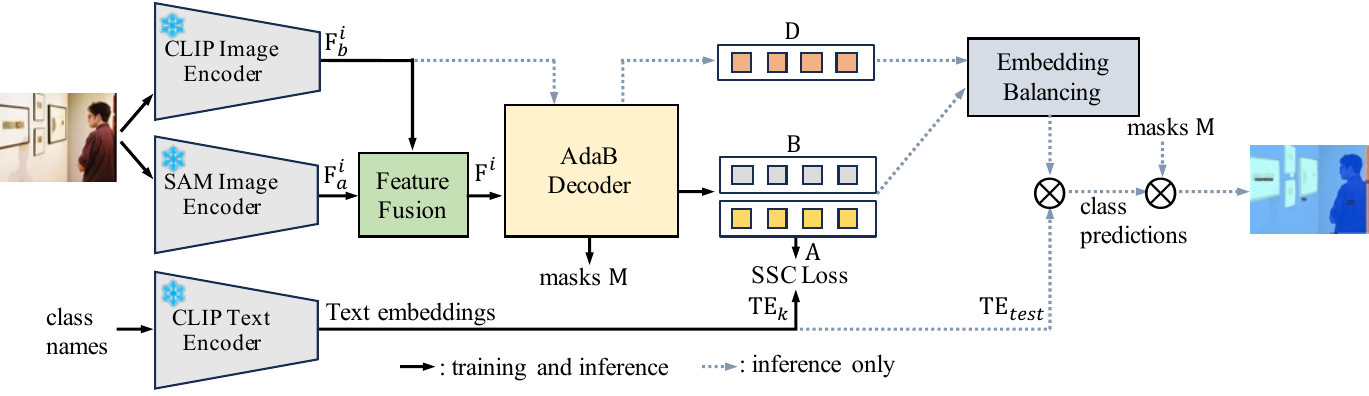}
    \caption{The architecture of our model EBSeg. We first obtain image features from two frozen image encoders and fuse them in a feature fusion module. After that, the fused features are input into our AdaB Decoder, which outputs masks $\mathbf{M}$ and image embeddings (including mask attention embeddings $\mathbf{B}$, fully supervised embeddings $\mathbf{A}$ and frozen embeddings $\mathbf{D}$). During training, we apply the SSC Loss to learn a consistent semantic structure from CLIP. During inference, we adaptively balance the three embeddings output by AdaB Decoder and obtain semantic segmentation results with the masks, balanced image embeddings, and text embeddings.}
    \label{fig:model architecture}
\end{figure*}

\textbf{CLIP and its transfer learning on downstream tasks.}  
CLIP \cite{radford2021learning} is proposed to align images and texts in a shared semantic space, enabling cross-modal understanding and transfer learning. It is trained on a large dataset of image-text pairs with contrastive loss to get a strong open-vocabulary recognition ability. CLIP can be directly used in the zero-shot image classification task.

After CLIP was released, many works have explored using it in various downstream tasks. For few-shot image classification, CoOp \cite{zhou2022learning} and CoCoOp \cite{zhou2022conditional} use prompt tuning \cite{zhong2021factual, li2021prefix, lester2021power} to adapt the text embeddings of CLIP to task-specific classes with relatively low cost.\
% For zero-shot multi-label classification, DualCoOp \cite{sun2022dualcoop} employs a dual prompt tuning method.
For fully supervised dense prediction, DenseCLIP \cite{rao2022denseclip} directly finetunes the CLIP model and proposes a vision-to-language prompting method to leverage the prior knowledge of image contexts. CLIP Surgery \cite{li2023clip} explores using surgery-like modifications for CLIP inference architecture and features, achieving good explainability and enhancement in multiple open-vocabulary tasks.

Our work explores how to fully leverage the powerful image-text alignment capability of CLIP in the challenging task of open-vocabulary semantic segmentation. 
% This task is challenging, since training on downstream semantic segmentation datasets often makes the model overfit to the classes in those datasets.

\noindent \textbf{Open-vocabulary semantic segmentation.}
Some early works \cite{zhao2017open, bucher2019zero, xian2019semantic} on this task focused on how to project image features and text features into a shared feature space, which is hard because images and texts are in two different modalities. Recently, benefiting from the powerful open-vocabulary recognition ability of CLIP, many works have attempted to apply the CLIP model on this task. \cite{ding2021decoupling, liang2023open, xu2022simple} adopt a two-stage framework, where a mask generator is leveraged to extract category-agnostic masks. Then the masks are used to get many crops of the input image and the crops will be fed into CLIP for mask classification results. MaskCLIP \cite{ding2023open} also uses a category-agnostic mask generator, but it does not use the masks to crop the input image. MaskCLIP \cite{ding2023open} proposes a Relative Mask Attention module where it uses the masks as attention masks in the self-attention layers of CLIP image encoder to get mask attention embeddings. ODISE \cite{xu2023open} explores using a frozen diffusion model \cite{rombach2022high} to extract image features as the input for a Mask2former \cite{cheng2022masked} head.
SAN \cite{xu2023side} adds a lightweight image encoder to get masks and attention masks corresponding to each mask. Like MaskCLIP \cite{ding2023open}, the attention masks are fed to the last few layers of CLIP to obtain mask attention embeddings for mask classification. 
% SAN \cite{xu2023side} follows the Relative Mask Attention module in MaskCLIP \cite{ding2023openvocabulary} and tries to add a light-weight ViT \cite{dosovitskiy2020image} image encoder to get masks and attention masks corresponding to each mask. 
Different from MaskCLIP \cite{ding2023open}, the attention masks in SAN \cite{xu2023side} are per-head, which means SAN \cite{xu2023side} produces different attention masks for each attention head in CLIP image encoder.

Our work explores how to overcome the overfitting challenge faced by CLIP-based methods. To achieve this goal, we design the AdaB Decoder and SSC Loss for better generalization on new classes. 
% What's more, we leverage a frozen SAM image encoder to complement the important spatial information of CLIP features.
\section{Method}
\label{sec:method}

\begin{figure*}
    \centering
    % \includesvg[width=0.8\linewidth]{figures/decoder.svg}
    \includegraphics[width=0.75\linewidth]{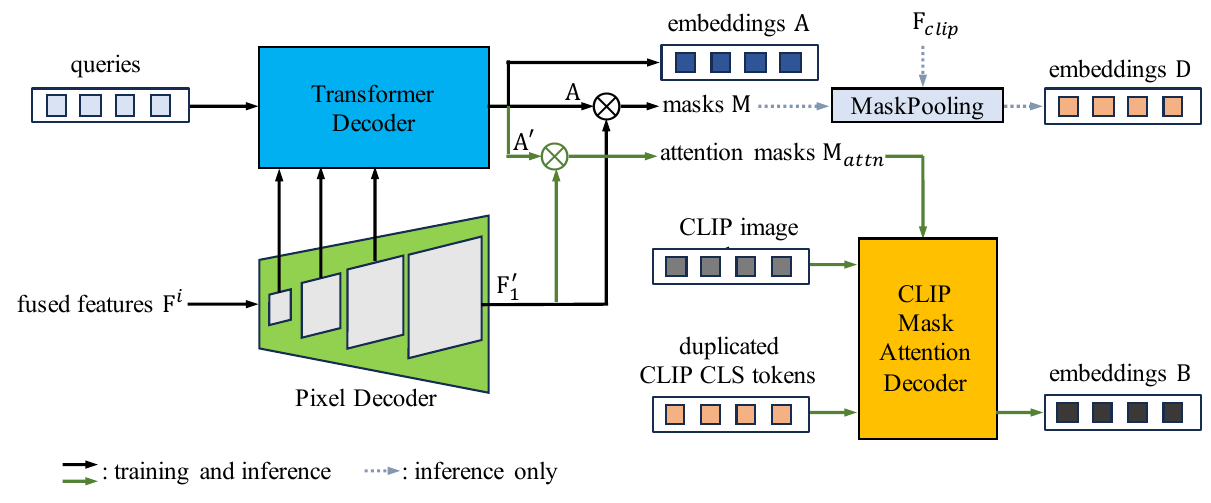}
    \caption{Detailed structure of AdaB Decoder. We first input fused image features into the Pixel Decoder. The outputs of the first three stages are then fed to the Transformer Decoder which outputs image embeddings $\mathbf{A}$ and $\mathbf{A}^{'}$. Then we obtain masks $\mathbf{M}$ with $\mathbf{A}$ and the largest feature map $\mathbf{F}_{1}^{'}$ from the Pixel Decoder. We obtain per-head attention masks $\mathbf{M}_{attn}$ with per-head embeddings $\mathbf{A}^{'}$ and $\mathbf{F}_{1}^{'}$. Finally, we perform masked self-attention in the last few blocks of CLIP image encoder with $\mathbf{M}_{attn}$ to get mask attention embeddings $\mathbf{B}$.}
    \label{fig:decoder}
\end{figure*}

\subsection{Method Overview}

In \cref{fig:model architecture}, we present the architecture of our open-vocabulary semantic segmentation model EBSeg. In this framework, we first input the image into both frozen CLIP and SAM image encoders. The obtained image features from the two image encoders are then fused in a fusion module (\cref{subsec:feature extract and fuse}). After that, the fused features are fed into our AdaB Decoder which outputs masks and image embeddings (including mask attention embeddings $\mathbf{B}$, fully supervised embeddings $\mathbf{A}$ and frozen embeddings $\mathbf{D}$) (\cref{subsec:decoder}). In our model, we apply the Semantic Structure Consistency loss (SSC Loss) to learn a consistent semantic structure from CLIP (\cref{subsec:loss}). During inference, we propose to adaptively balance the three different image embeddings to fully exploit their ability to recognize training classes and the excellent generalization ability for new classes (\cref{subsec:inference}). Finally, we obtain semantic segmentation results with the masks, balanced image embeddings, and text embeddings.

\subsection{Image Feature Extraction and Fusion}
\label{subsec:feature extract and fuse}
As mentioned before, the feature maps of CLIP lack important spatial information for semantic segmentation. Therefore, we propose to utilize a frozen SAM image encoder to complement the spatial information.
% Below, we explain how to do this.

Given an image $\mathbf{I}\in \mathbb{R}^{H \times W \times 3}$, we input it into the SAM image encoder and get image features $\mathbf{F}_{a}^{i} \in \mathbb{R}^{h_a \times w_a \times C_a} (i=1,2,3)$ from the last three global attention blocks. Meanwhile, we downsample $\mathbf{I}$ to $\mathbf{I}^{'} \in \mathbb{R}^{\frac{H}{p} \times \frac{W}{p} \times 3}$ (where $p$ is the downsample ratio). We then feed $\mathbf{I}^{'}$ into CLIP image encoder to get image features $\mathbf{F}_{b}^{i} \in \mathbb{R}^{h_b \times w_b \times C_b} (i=1,2,3)$ from the number $L/4, L/2,$ and $3L/4$ blocks of CLIP ($L$ is the total number of blocks in CLIP image encoder; $L=12$ for CLIP ViT-B and $L=24$ for CLIP ViT-L).

We employ a simple addition approach to fuse the image features from two frozen models. Firstly, we use a linear layer to match the channel number of $\mathbf{F}_{b}$ to that of $\mathbf{F}_{a}$. Then, we upsample or downsample $\mathbf{F}_{a}$ and $\mathbf{F}_{b}$ to $\mathbf{F}_{v}^{i} \in \mathbb{R}^{\frac{H}{s} \times \frac{W}{s} \times C_{a}} (v=a,b; i=1,2,3; s=2^{i+2})$. Finally we perform element-wise addition with $\mathbf{F}_{a}$ and $\mathbf{F}_{b}$, obtaining fused features $\mathbf{F}^{i} \in \mathbb{R}^{\frac{H}{s} \times \frac{W}{s} \times C_a} (i=1,2,3; s=2^{i+2})$:

\begin{equation}
  \mathbf{F}^{i} = \mathbf{F}_{a}^i + \mathbf{F}_{b}^{i}.
  \label{eq:add fusion}
\end{equation}

\subsection{AdaB Decoder}
\label{subsec:decoder}
In this section, we show the detailed architecture of our AdaB Decoder (\cref{fig:decoder}). Note that we will present how to adaptively balance the image embeddings later in \cref{subsec:inference}.

The AdaB Decoder consists of three components: Pixel Decoder, Transformer Decoder, and CLIP Mask Attention Decoder. The Pixel Decoder and Transformer Decoder follow Mask2former \cite{cheng2022masked}, and the CLIP Mask Attention Decoder follows the mask attention module in SAN \cite{xu2023side}.

Similar to Mask2former \cite{cheng2022masked}, we input the fused multi-scale image features $\mathbf{F}^{i}$ into the Pixel Decoder. The outputs of the first three stages in Pixel Decoder are then fed into the Transformer Decoder, from which we get the fully supervised mask image embeddings $\mathbf{A} \in \mathbb{R}^{N \times C}$ ($N$ is the number of queries of the Transformer Decoder). We then perform matrix multiplication between $\mathbf{A}$ and the largest feature map $\mathbf{F}^{'}_{1} \in \mathbb{R}^{\frac{H}{4} \times \frac{W}{4} \times C}$ from Pixel Decoder to obtain masks $\mathbf{M} \in \mathbb{R}^{\frac{H}{4} \times \frac{W}{4} \times N}$: 
\begin{equation}
  \mathbf{M} = \mathbf{F}^{'}_{1} \times \mathbf{A}^T.
  \label{eq:mask}
\end{equation}

Furthermore, we increase the channel numbers of $\mathbf{A}$ to $C \times n$ using a linear layer ($n$ is the number of heads in the multi-head self-attention layers of CLIP image encoder), getting per-head image embeddings $\mathbf{A}^{'} \in \mathbb{R}^{N \times (C \times n)}$. We then perform matrix multiplication between $\mathbf{A}^{'}$ and $\mathbf{F}^{'}_{1}$, obtaining per-head attention masks $\mathbf{M}_{attn} \in \mathbb{R}^{\frac{H}{4} \times \frac{W}{4} \times n \times N}$ which will be used in the CLIP Mask Attention Decoder.

In the CLIP Mask Attention Decoder, we use the last $l$ ($l=\frac{L}{4}$) transformer blocks from CLIP image encoder. Firstly, we get the output tokens $\mathbf{U}$ from the last ($l+1$)th block of CLIP image encoder. Then we duplicate the CLS token of $\mathbf{U}$ $N$ times as queries for the CLIP Mask Attention Decoder. Then, we concatenate the queries and $\mathbf{U}$ together and input them into the CLIP Mask Attention Decoder, with $\mathbf{M}_{attn}$ as the per-head attention masks for the multi-head self-attention layers of those $l$ CLIP transformer blocks. The output of CLIP Mask Attention Decoder is mask attention embeddings $\mathbf{B} \in \mathbb{R}^{N \times C}$.

During inference, to improve the recognition ability for new classes that do not exist in the training set, we perform mask pooling with $\mathbf{M}$ and the final output $\mathbf{F}_{clip} \in \mathbb{R}^{\frac{H}{16} \times \frac{W}{16} \times C}$ of CLIP image encoder to get frozen image embeddings $\mathbf{D} \in \mathbb{R}^{N \times C}$:
\begin{equation}
  \mathbf{D} = MaskPooling(\mathbf{M}, \mathbf{F}_{clip}).
  \label{eq:mask_pooling}
\end{equation}
% $\mathbf{D}$ has strong generalization ability for the new classes, because it's extracted from the original feature map of CLIP.

\subsection{SSC Loss}
\label{subsec:loss}

% \begin{figure}
%     \centering
%     \includesvg[width=1\linewidth]{figures/loss.svg}
%     \caption{Left: The CLIP text embeddings in the embedding space. The yellow circles with numbers are corresponding to some classes in the training set, while the blue ones are corresponding to some classes that don’t exist in the training set. Middle: The image embeddings in the embedding space after training with only cross-entropy loss. The yellow circles with solid boundary are training classes’ corresponding image embeddings, the yellow circles with dashed boundary are training classes’ text embeddings, and the gray circles are some test classes’ corresponding image embeddings. With only cross-entropy loss, the training image embeddings will get closer and closer to their corresponding text embeddings (the arrows represents the optimization directions of image embeddings) during training, which is good. However, this optimizing process does not consider the relationships between image embeddings. Right: With the image embedding relationship loss, during training, the image embedding of a specific category will not only get closer to its corresponding text embedding but also have a more similar relationship with the other classes’ image embeddings to that of all training text embeddings in the feature space.}
%     \label{fig:embedding loss}
% \end{figure}
In this section, we will introduce the Semantic Structure Consistency Loss (SSC Loss).
Our SSC Loss draws inspiration from unsupervised segmentation methods like STEGO \cite{hamilton2021unsupervised}.
Works like STEGO \cite{hamilton2021unsupervised} mainly focus on the unsupervised segmentation task and
usually aim to distill image feature semantic similarity from a pretrained image encoder.
Designed for OVSS, our SSC loss is proposed to enhance generalization ability for new classes by distilling semantic similarity from text features output by frozen CLIP text encoder to image features.
% This loss constrains distribution of the image embeddings to be similar with the distribution of text embeddings during training. 
By introducing the SSC Loss, our model learns more about the latent knowledge from the CLIP feature space. Thus our model can learn a consistent semantic structure from CLIP and gain a stronger generalization ability for new classes that do not exist in the training set.

During training, assuming that an image has $k$ ground truth masks and classes, after Hungarian matching, we match these $k$ ground truth masks with $k$ prediction masks (from $\mathbf{M}$) and $k$ image embeddings $\mathbf{IE}_k \in \mathbb{R}^{k \times C}$ (from $\mathbf{A}$).

After that, we calculate the cosine similarities $\mathbf{CS}_{text}^{i,j} \in \mathbb{R}^{1}$ between the text embeddings $\mathbf{TE}_k \in \mathbb{R}^{k \times C}$ (produced by the CLIP text encoder) of the $k$ ground truth classes:
\begin{equation}
  \mathbf{CS}_{text}^{i,j} = \frac{\mathbf{TE}_k^i \cdot \mathbf{TE}_k^j}{|\mathbf{TE}_k^i| * |\mathbf{TE}_k^j|},
  \label{eq:cs_1}
\end{equation}
where $\mathbf{TE}_k^i$, $\mathbf{TE}_k^j \in \mathbb{R}^{1 \times C}$ from $\mathbf{TE}_k$, $i$, $j \leq k$, and $\cdot$ denotes dot product. Then, we calculate the cosine similarities $\mathbf{CS}_{image}^{i,j} \in \mathbb{R}^{1}$ between the matched $k$ image embeddings $\mathbf{IE}_k \in \mathbb{R}^{k \times C}$ (from $\mathbf{A} \in \mathbb{R}^{N \times C}$):
\begin{equation}
  \mathbf{CS}_{image}^{i,j} = \frac{\mathbf{IE}_k^i \cdot \mathbf{IE}_k^j}{|\mathbf{IE}_k^i| * |\mathbf{IE}_k^j|},
  \label{eq:cs_2}
\end{equation}
where $\mathbf{IE}_k^i$, $\mathbf{IE}_k^j \in \mathbb{R}^{1 \times C}$ from $\mathbf{IE}_k$ and $i$, $j \leq k$. 

Finally, we compute the $L2$ distance between $\mathbf{CS}_{image}^{i,j}$ and $\mathbf{CS}_{text}^{i,j}$ as the SSC Loss:
\begin{equation}
  L_{SSC} = \frac{1}{k^2} \sum_{i=1}^{k} \sum_{j=1}^{k} \|\mathbf{CS}_{text}^{i,j} - \mathbf{CS}_{image}^{i,j}\|_2.
  \label{eq:relationship loss}
\end{equation}

During training, we apply the semantic segmentation loss in Mask2former \cite{cheng2022masked}. Our total loss during training is:
\begin{equation}
  L_{total} = L_{sem\_seg} + \lambda L_{SSC}.
  \label{eq:total loss}
\end{equation}

\subsection{Adaptively Balancing and Inference}
\label{subsec:inference}
In this section, we show how to adaptively balance the image embeddings (mask attention embeddings $\mathbf{B}$, fully supervised embeddings $\mathbf{A}$ and frozen embeddings $\mathbf{D}$) for better recognition ability on both training and new classes.

During inference, we assume that there are $K$ classes $C_{test}$ in the test set, where $f$ ($f<K$) classes exist in the training classes. Firstly, we use the CLIP text encoder to extract text embeddings $\mathbf{TE}_{test} \in \mathbb{R}^{K \times C}$ for $C_{test}$:
\begin{equation}
  \mathbf{TE}_{test} = \Theta (C_{test})
  \label{eq:text_embeddings},
\end{equation}
where $\Theta$ denotes the CLIP Text Encoder.

Then, we adaptively balance image embeddings ($\mathbf{A}$, $\mathbf{B}$ and $\mathbf{D} \in \mathbb{R}^{N \times C}$) with weights ($\alpha$, $\beta$ and $\gamma$) for both training and new classes:
% \begin{equation}
%   \mathbf{E}_{train} = \mathbf{A}^{\alpha} \odot \mathbf{B}^{\beta} \odot \mathbf{D}^{1-\alpha-\beta},
%   \label{eq:mean_embed_1}
% \end{equation}
% \begin{equation}
%   \mathbf{E}_{new} = \mathbf{A}^{\gamma} \odot \mathbf{B}^{\beta} \odot \mathbf{D}^{1-\gamma-\beta},
%   \label{eq:mean_embed_2}
% \end{equation}
\begin{equation}
  \mathbf{E}_{train} = \alpha*\mathbf{A} + \beta*\mathbf{B} + (1-\alpha-\beta)*\mathbf{D},
  \label{eq:mean_embed_1}
\end{equation}
\begin{equation}
  \mathbf{E}_{new} = \gamma*\mathbf{A} + \beta*\mathbf{B} + (1-\gamma-\beta)*\mathbf{D},
  \label{eq:mean_embed_2}
\end{equation}
where $\mathbf{E}_{train}$, $\mathbf{E}_{new} \in \mathbb{R}^{N \times C}$ are balanced embeddings for training and new classes respectively; $\odot$ denotes element-wise multiplication.
%关于平均值的说明
%请注意，在上面的公式中，我们展示了使用算数平均值来获取balanced embeddings，但是我们发现几何平均值更好，精度更高，具体描述见补充材料
In \cref{eq:mean_embed_1} and \cref{eq:mean_embed_2}, we use arithmetic mean to obtain balanced embeddings. However, we find that using geometric mean achieves 
slightly higher accuracy. Please refer to the supplementary materials for more details and experiments about this.

Next, we get mask classification predictions for both training and new classes:
\begin{equation}
  \mathbf{P}_{train} = \mathbf{E}_{train} \times \mathbf{TE}_{train}^T,
  \label{eq:logit_1}
\end{equation}
\begin{equation}
  \mathbf{P}_{new} = \mathbf{E}_{new} \times \mathbf{TE}_{new}^T,
  \label{eq:logit_2}
\end{equation}
where $\mathbf{TE}_{train} \in \mathbb{R}^{f \times C}$ and $\mathbf{TE}_{new} \in \mathbb{R}^{(K-f) \times C}$ from $\mathbf{TE}_{test}$, $\mathbf{P}_{train} \in \mathbb{R}^{N \times f}$ and $\mathbf{P}_{new} \in \mathbb{R}^{N \times (K-f)}$. The mask classification predictions $\mathbf{P} \in \mathbb{R}^{N \times K}$ for all test classes are:
\begin{equation}
  \mathbf{P} = Concatenate(\mathbf{P}_{train},\mathbf{P}_{new}).
  \label{eq:logit}
\end{equation}
Please note that we rearrange the order of $\mathbf{P}$ along the second dimension to match the order of the ground truth labels.

Finally, we obtain the semantic segmentation results $\mathbf{S} \in \mathbb{R}^{\frac{H}{4} \times \frac{W}{4} \times K}$:
\begin{equation}
  \mathbf{S}=\mathbf{M} \times \mathbf{P}.
  \label{eq:sem_seg}
\end{equation}

Note that "adaptively" means using different preset balancing weights specific for training and test classes to achieve better open vocabulary recognition ability, rather than automatically setting balancing weights. We will try using learnable balancing weights in our future work.
\section{Experiments}
\label{sec:experiments}
% In this section, we first introduce our experiment setup, which includes datasets, evaluation metric and implement details. Then we compare our method with other methods on several datasets. Lastly, we do ablation studies to show the effectiveness of our method.
\begin{table*}[htpb]
  \centering
  \begin{tabular}{c|c|c|c|c|c|c|c}
    \toprule
    Method & VLM & Training Dataset & A-847 & PC-459 & A-150 & PC-59 & VOC \\
    \midrule
    LSeg+ \cite{li2022languagedriven} & ALIGN RN-101 & COCO-Stuff & 2.5 & 5.2 & 13.0 & 36.0 & 59.0\\ 
    OpenSeg \cite{ghiasi2022scaling} & ALIGN RN-101 & COCO Panoptic\cite{kirillov2019panoptic1} & 4.0 & 6.5 & 15.3 & 36.9 & 60.0\\
    \midrule
    LSeg+ \cite{li2022languagedriven} & ALIGN EN-B7 & COCO-Stuff & 3.8 & 7.8 & 18.0 & 46.5 & -\\ 
    OpenSeg \cite{ghiasi2022scaling} & ALIGN EN-B7 & COCO Panoptic\cite{kirillov2019panoptic1}+Loc.Narr. & 8.8 & 12.2 & 28.6 & 48.2 & 72.2\\
    \midrule
    ZegFormer \cite{ding2021decoupling} & CLIP ViT-B/16 & COCO-Stuff & 5.6 & 10.4 & 18.0 & 45.5 & 89.5\\
    SimSeg \cite{xu2022simple} & CLIP ViT-B/16 & COCO-Stuff & 6.9 & 9.7 & 21.1 & 51.9 & 91.8\\
    OVSeg \cite{liang2023open} & CLIP ViT-B/16 & COCO-Stuff+COCO Caption\cite{chen2015microsoft} & 7.1 & 11.0 & 24.8 & 53.3 & 92.6\\
    SAN \cite{xu2023side} & CLIP ViT-B/16 & COCO-Stuff & 10.1 & 12.6 & 27.5 & 53.8 & 94.0\\
    % CAT-Seg & CLIP ViT-B/16 & COCO-Stuff & 8.4 & 16.6 & 27.2 & 57.5 & 93.7\\
    EBSeg(ours) & CLIP ViT-B/16 & COCO-Stuff & \textbf{11.1} & \textbf{17.3} & \textbf{30.0} & \textbf{56.7} & \textbf{94.6}\\
    % ours & CLIP ViT-B/16 & COCO-Stuff & \textbf{11.3} & \textbf{17.9} & \textbf{29.2} & \textbf{57.6} & \textbf{95.0}\\
    % ours & CLIP ViT-B/16 & COCO-Stuff & \textbf{11.5} & \textbf{18.1} & \textbf{30.1} & \textbf{58.0} & \textbf{95.0}\\
    \midrule
    SimSeg \cite{xu2022simple} & CLIP ViT-L/14 & COCO-Stuff & 7.1 & 10.2 & 21.7 & 52.2 & 92.3\\
    MaskCLIP \cite{ding2023open} & CLIP ViT-L/14 & COCO Panoptic & 8.2 & 10.0 & 23.7 & 45.9 & -\\
    OVSeg \cite{liang2023open} & CLIP ViT-L/14 & COCO-Stuff & 9.0 & 12.4 & 29.6 & 55.7 & 94.5\\
    ODISE \cite{xu2023open} & SD+CLIP ViT-L/14 & COCO Panoptic & 11.1 & 14.5 & 29.9 & 57.3 & -\\
    SAN \cite{xu2023side} & CLIP ViT-L/14 & COCO-Stuff & 12.4 & 15.7 & 32.1 & 57.7 & 94.6\\
    % CAT-Seg & CLIP ViT-L/14 & COCO-Stuff & 10.8 & 20.4 & 31.5 & 62.0 & 96.6\\
    EBSeg(ours) & CLIP ViT-L/14 & COCO-Stuff & \textbf{13.7} & \textbf{21.0} & \textbf{32.8} & \textbf{60.2} & \textbf{96.4}\\
    \bottomrule
  \end{tabular}
  \caption{Comparison with state-of-the-art methods. We use mIoU as the evaluation metric. VLM denotes vision-language model. ALIGN \cite{jia2021scaling} is a vision-language model. EN-B7 \cite{tan2019efficientnet} is the image backbone used by ALIGN \cite{jia2021scaling}. Loc.Narr. stands for Localized Narrative \cite{pont2020connecting}, which contains detailed natural language descriptions for multiple datasets. SD denotes the stable diffusion model \cite{rombach2022high}.}
  \label{tab:main table}
\end{table*}

\subsection{Experiment Setup}
\textbf{Datasets and evaluation protocol.} We train our model on the COCO-Stuff \cite{caesar2018coco} dataset, and evaluate the model on five other benchmarks, including Pascal VOC \cite{everingham2012pascal}, Pascal Context-59 (PC-59) \cite{mottaghi2014role}, Pascal Context-459 (PC-459) \cite{mottaghi2014role}, ADE20K-150 (A-150) \cite{zhou2017scene}, ADE20K-847 (A-847)\cite{zhou2017scene}. There are 171 densely annotated classes in the COCO-stuff dataset, which contains 118K training images, 5K validation images, and 41K test images. We train our model on the training set of COCO-Stuff. Pascal VOC has 20 classes, with 1464 training and 1449 validation images. Pascal Context has 5K training images and 5K validation images, with two types of annotations: 59 classes annotated in Pascal Context-59 and 459 classes annotated in Pascal Context-459. ADE20K contains 20K training images and 2K validation images, and it has two sets of annotated classes: ADE20K-150 with 150 classes and ADE20K-847 with 847 classes. Same as early works, we adopt mean Intersection over Union (mIoU) as the evaluation metric for all our experiments.

\noindent \textbf{Implement details.} We use OpenAI pretrained CLIP model \cite{radford2021learning} in our experiments, including a ViT-B/16 model and a ViT-L/14 model. The ViT-B model of SAM \cite{kirillov2023segment} is used for all our experiments. The input image resolution is $640^{2}$, and the downsample ratio $p$ for the CLIP image encoder is set to $0.5$ and $0.7$ for our ViT-B and ViT-L models respectively. For the CLIP and SAM image encoders, we freeze all their parameters except for their positional embeddings. For the AdaB Decoder, the Transformer Decoder in it has 9 layers and 100 queries. The hidden dimension of the Transformer Decoder is 256, and its output dimension $C$ is set to the same as the dimension of CLIP features (512 for CLIP ViT-B and 640 for CLIP ViT-L). For adaptively image embedding balancing, we set $\alpha=0.2$, $\beta=0.7$ and $\gamma=0$ by default. We use CLIP Surgery \cite{li2023clip} to get better CLIP final output $\mathbf{F}_{clip}$ for frozen embeddings $\mathbf{D}$. For SSC loss, the loss weight $\lambda$ is set to 10. An auxiliary loss (the loss $L_{sem\_seg}$) is added to every intermediate Transformer Decoder layer. Note that we only apply our SSC Loss to fully supervised embeddings $A$ from the last layer of the Transformer Decoder. 
% and the $L_{cls}$ is also applied to mask attention embeddings $\mathbf{B}$ corresponding to attention masks output by every layer of the Transformer Decoder
Our models are implemented with Pytorch \cite{paszke2019pytorch} and Detectron2 \cite{wu2019detectron2}. AdamW \cite{loshchilov2017decoupled} optimizer is used with the initial learning rate of $1 \cdot 10^{-4}$, weight decay of $5 \cdot 10^{-2}$. We set the batch size to $16$, and train models for $120k$ iterations.

\subsection{Comparison with State-of-the-Art methods}

\begin{table}[]
    \centering
    \begin{tabular}{c|c|c|c}
     \toprule
     AIB & AdaB Decoder & SSC Loss & mIoU \\
     \midrule
       % None&  &  & 24.5  \\
       Swin-B \cite{liu2021swin} &  &  & 27.9  \\
       SAM-B \cite{kirillov2023segment}&  &  & 28.5  \\
       SAM-B \cite{kirillov2023segment}& \checkmark &  & 29.2  \\
       SAM-B \cite{kirillov2023segment}& & \checkmark & 29.0  \\
       SAM-B \cite{kirillov2023segment}& \checkmark & \checkmark & \textbf{30.0}  \\
     \bottomrule
    \end{tabular}
    \caption{The ablation study on the proposed components. SAM-B denotes the ViT-B model of SAM \cite{kirillov2023segment}}
    \label{tab:component analisis}
\end{table}

In \cref{tab:main table} we compared our method EBSeg with other methods on several datasets. We list the datasets and vision-language models (VLM) used in various methods in the table. To ensure a fair comparison, we group the results using the same vision-language model together.

Overall, compared with other methods using CLIP ViT models, our method outperforms the best of them across all test datasets. With CLIP ViT-B/16 as the vision-language model, our method gains +1.0\% mIoU, + 4.7\% mIoU, +2.5\% mIoU, +2.9\% mIoU, +0.6\% mIoU improvements on A-847, PC-459, A-150, PC-59, VOC, respectively. When using CLIP ViT-L, the improvements are +1.3\% mIoU, + 5.3\% mIoU, +0.7\% mIoU, +2.5\% mIoU, +1.8\% mIoU on the five datasets respectively. These results demonstrate the effectiveness of our proposed method EBSeg.

We also show some visualization results of our method on the ADE20K-150 validation set in \cref{fig:visualization}. More visualization results can be found in the supplementary materials.

\begin{table}[]
    \centering
    \begin{tabular}{m{3cm}|c|c}
     \toprule
     type(s) of embeddings \centering & embeddings  & mIoU \\
     \midrule
       & $\mathbf{A}$ & 17.7 \\
      one \centering & $\mathbf{B}$ & 29.0 \\
       & $\mathbf{D}$ &  19.7 \\
     \midrule
       & $\mathbf{A}$ and $\mathbf{B}$ & 29.4 \\
      two \centering & $\mathbf{A}$ and $\mathbf{D}$ & 26.3 \\
       & $\mathbf{B}$ and $\mathbf{D}$ & 29.7 \\
     \midrule
      three \centering & $\mathbf{A}$, $\mathbf{B}$ and $\mathbf{D}$ & \textbf{30.0}  \\
     \bottomrule
    \end{tabular}
    \caption{The ablation study on the Adaptively Balanced Decoder (AdaB Decoder).}
    \label{tab:embedding conbination}
\end{table}

\begin{table}[]
    \centering
    \begin{tabular}{c|c|c|c}
     \toprule
     AdaB Decoder & mIoU train & mIoU new & mIoU\\
     \midrule
       $\times$ & 39.5 & 20.5 & 29.0  \\
       \checkmark & 40.3(+0.8) & 21.6(+1.1) & 30.0(+1.0)  \\
     \bottomrule
    \end{tabular}
    \caption{The ablation study on the influence of the proposed AdaB Decoder on the mIoU of the training (mIoU training) and new (mIoU new) classes in the ADE20K-150 validation set. In the second row, we only use the mask attention embeddings $\mathbf{B}$ and in the third row we adaptively balance the image embeddings $\mathbf{A}$, $\mathbf{B}$, and $\mathbf{D}$. Note that ADE20K-150 has 67 classes that exist in the COCO-Stuff dataset and 83 classes that do not exist in COCO-Stuff.}
    \label{tab:train and new combine}
\end{table}

\subsection{Ablation Studies}
We conduct ablation studies on the ADE20K-150 dataset. In this section, all models use CLIP ViT-B/16 as the vision-language model by default.

\noindent \textbf{Component Analysis.} We conduct ablation studies in \cref{tab:component analisis} to analyze the effect of the essential components of our method.
% The first row in \cref{tab:component analisis} represents a model that only uses a CLIP image encoder as backbone, and leverages mask attention embeddings $\mathbf{B}$ to compute semantic segmentation results for all classes. After we use a frozen Swin-B \cite{liu2021swin} model as the additional image backbone and fuse its image features with the CLIP features, the mIoU is improved by 3.4\%.
The first row in \cref{tab:component analisis} represents a model that uses a CLIP image encoder and a frozen Swin-B \cite{liu2021swin} as backbones, and leverages mask attention embeddings $\mathbf{B}$ to compute semantic segmentation results for all classes.
If we replace the frozen Swin-B with a frozen SAM-B \cite{kirillov2023segment} image encoder, the performance is further improved by 0.6\% mIoU. The AdaB Decoder improves the mIoU further by 0.7\% mIoU. With the SSC Loss, the mIoU improves further by 0.8\%. The experiments show that the three methods we propose are all effective for the open-vocabulary semantic segmentation task.

\begin{table}[]
    \centering
    \begin{tabular}{m{3cm}|m{2.2cm}|c}
     \toprule
     embeddings using SSC Loss \centering & layers using SSC Loss \centering & mIoU \\
     \midrule
       None \centering & None \centering & 29.2  \\
     \midrule
       & all layers \centering & 29.0 \\
       $\mathbf{A}$ \centering & last 3 layers \centering & 29.6 \\
       & last 1 layer \centering & \textbf{30.0} \\
     \midrule
        & all layers \centering & 28.6 \\
       $\mathbf{B}$ \centering & last 3 layers \centering & 28.8 \\
        & last 1 layer \centering & 29.1 \\
     \midrule
        & all layers \centering & 28.8 \\
       both $\mathbf{A}$ and $\mathbf{B}$ \centering & last 3 layers \centering & 29.3  \\
        & last 1 layer \centering & 29.7 \\
     \bottomrule
    \end{tabular}
    \caption{The ablation study on the Semantic Structure Consistency loss (SSC Loss). We apply the SSC Loss to different embeddings and different layers in our Adab Decoder during training.}
    %Note that our decoder use the deep supervision technique following Mask2former \cite{cheng2022masked}, so we actually have 10 layers to supervise.
    \label{tab:image embedding relationship loss}
\end{table}

\begin{table}[]
    \centering
    \begin{tabular}{c|c|c|c}
     \toprule
     SSC Loss & mIoU train & mIoU new & mIoU \\
     \midrule
       $\times$ & 40.6 & 19.9 & 29.2  \\
       \checkmark & 40.3(-0.3) & 21.6(+1.7) & 30.0(+0.8)  \\
     \bottomrule
    \end{tabular}
    \caption{Ablation study on the influence of the SSC Loss to the mIoU of the training and new classes in the ADE20K-150 validation set.}
    \label{tab:train and new loss}
\end{table}

\begin{figure*}[!t]
  \centering
  \begin{tabular}{ccccc}
    \rotatebox{90}{Input Image} &
      \includegraphics[width=0.21\textwidth]{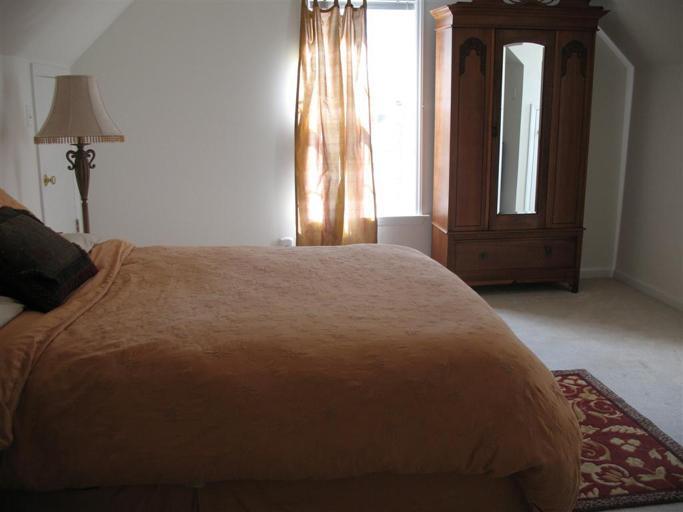} & \includegraphics[width=0.21\textwidth]{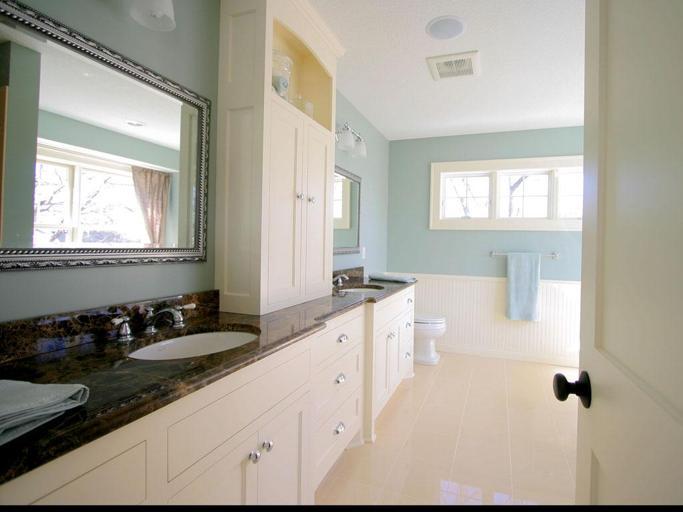} &
    \includegraphics[width=0.21\textwidth]{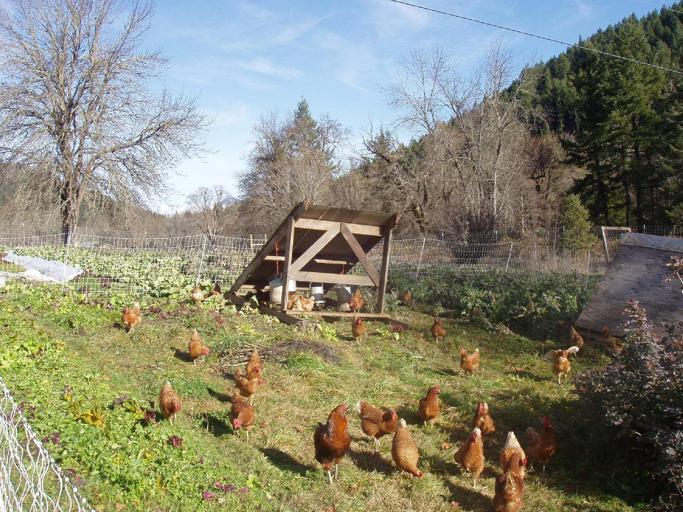} & \includegraphics[width=0.21\textwidth]{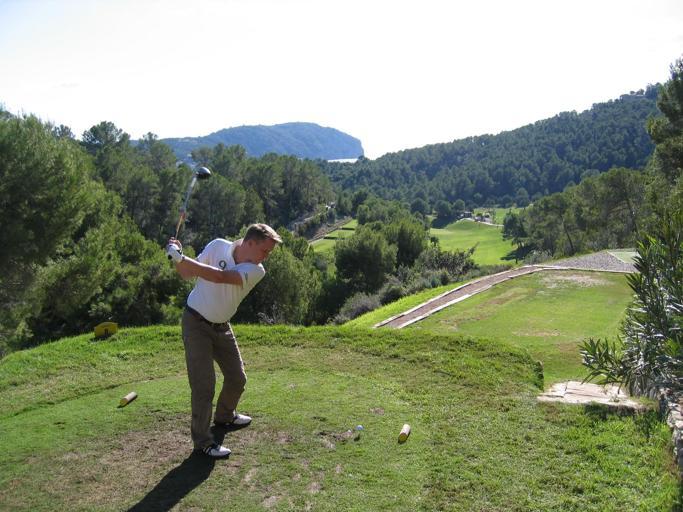} \\
  \rotatebox{90}{Ground Truth} &
  \includegraphics[width=0.21\textwidth]{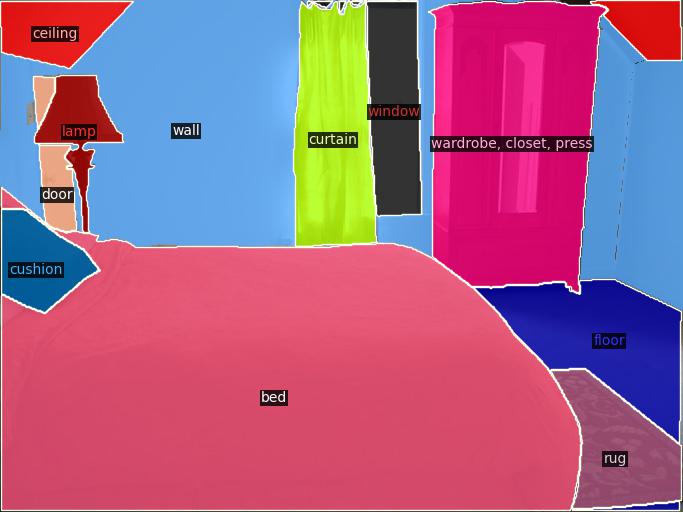} & \includegraphics[width=0.21\textwidth]{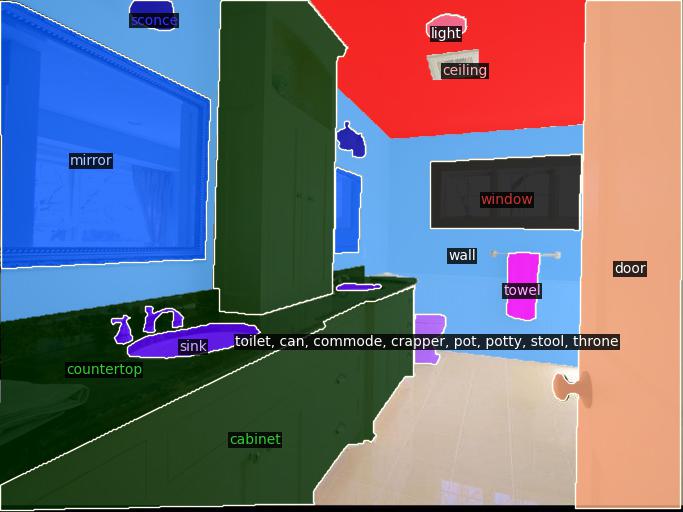} &
    \includegraphics[width=0.21\textwidth]{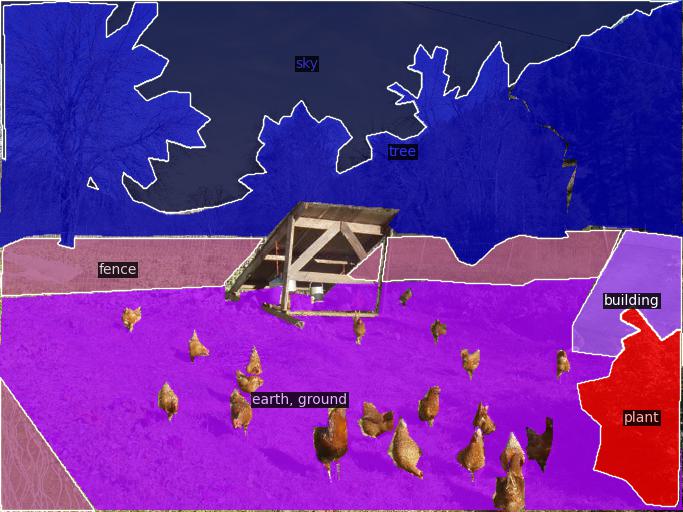} & \includegraphics[width=0.21\textwidth]{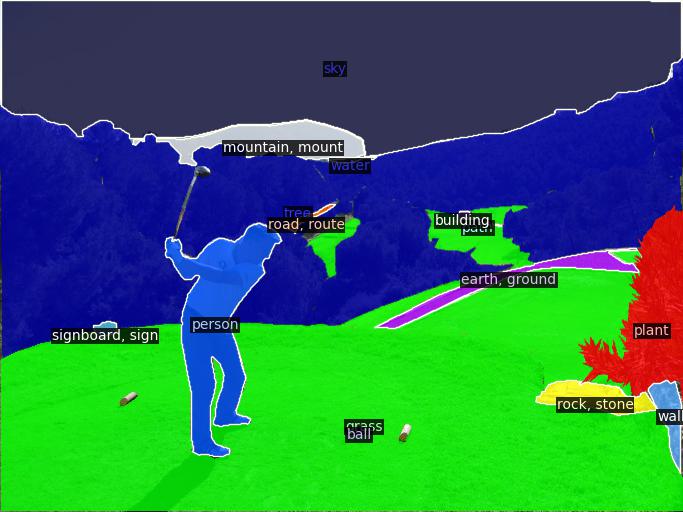} \\
\rotatebox{90}{Prediction} &
  \includegraphics[width=0.21\textwidth]{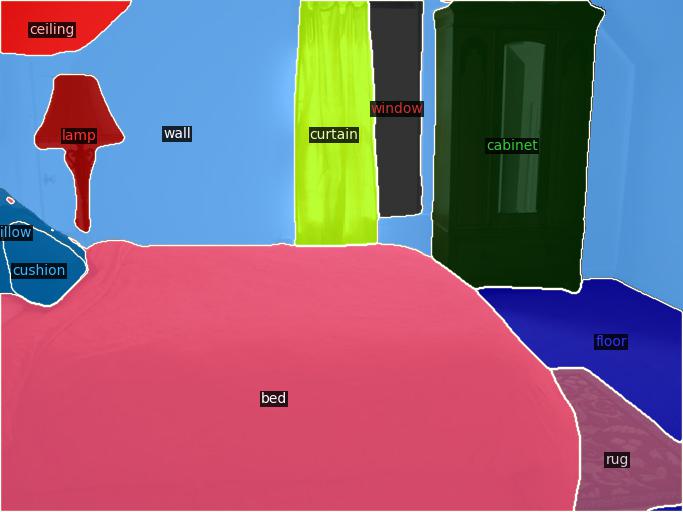} & \includegraphics[width=0.21\textwidth]{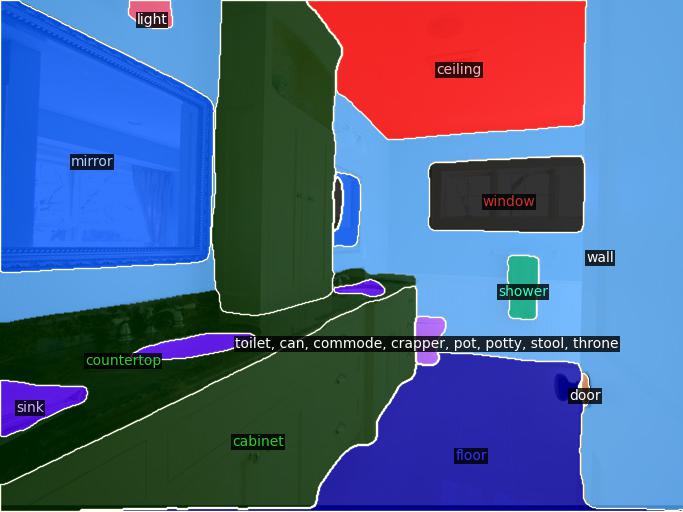} &
    \includegraphics[width=0.21\textwidth]{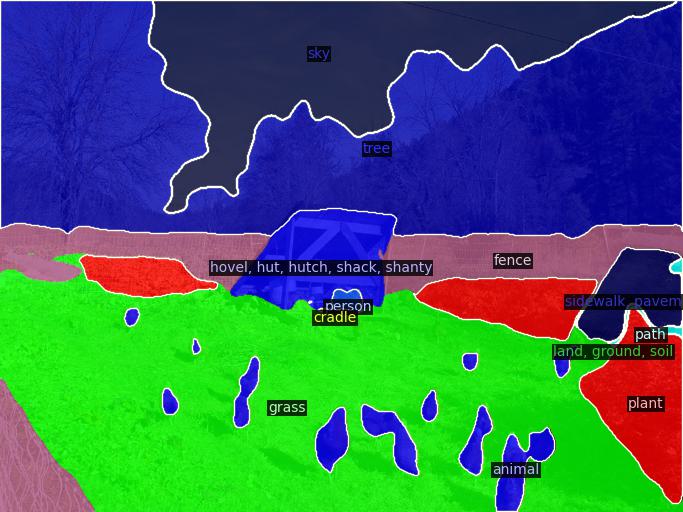} & \includegraphics[width=0.21\textwidth]{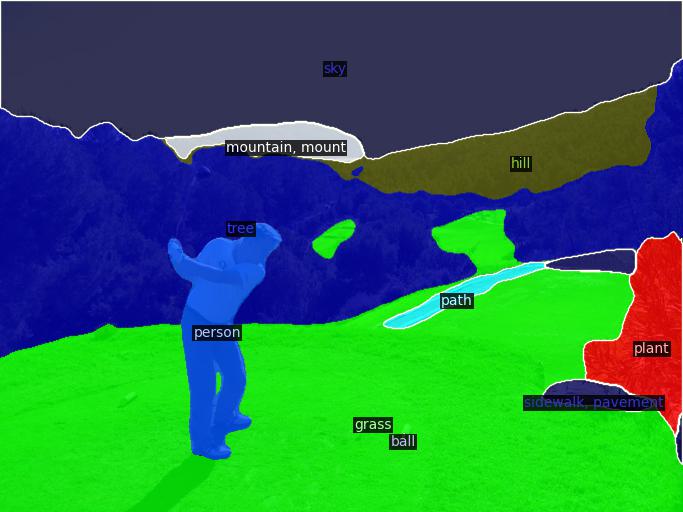} \\
  \end{tabular}
  \caption{Visualization examples of our model on ADE20K-150 validation set.}
  \label{fig:visualization}
\end{figure*}

\noindent \textbf{AdaB Decoder.}
In \cref{tab:embedding conbination}, we present the performances when we balance different image embeddings during inference. When using a single type of embeddings, the mIoU for $\mathbf{A}$, $\mathbf{B}$ and $\mathbf{D}$ embeddings are 17.7\%, 29.0\% and 19.7\% respectively. When balancing two types of embeddings, mIoU improves. The highest mIoU is achieved when we adaptively balance all image embeddings, which is 30.0\%.

In \cref{tab:train and new combine}, we show how AdaB Decoder influences mIoU of the training and new classes. After applying the AdaB Decoder, the mIoU of both training and new classes improves by a large margin. The improvement shows that with our AdaB Decoder, the model performs better at both training and new classes.

\noindent \textbf{SSC Loss.}
In \cref{tab:image embedding relationship loss}, we show the importance of the SSC Loss (Semantic Structure Consistency loss). Without the SSC Loss, the performance will drop from 30.0\% mIoU to 29.2\% mIoU. When we apply the loss to image embeddings $\mathbf{A}$ from all layers of the Transformer Decoder in our AdaB Decoder, the performance drops. Since the image embeddings from the first few layers lack rich semantic information, we do not apply the SSC Loss to the first few layers. When we apply the SSC Loss to $\mathbf{A}$ from the last 3 layers, the mIoU is improved from 29.2\% to 29.6\%. After we only apply the SSC Loss to embeddings $\mathbf{A}$ generated by the last layer, the mIoU is further improved to 30.0\%. 
We also find that when applying this loss only to embeddings $\mathbf{B}$ or both embeddings $\mathbf{A}$ and $\mathbf{B}$, the model's performance is unsatisfactory. We believe this is because the process of obtaining mask attention embeddings $\mathbf{B}$ involves many frozen parameters, which makes it difficult to optimize the SSC Loss.

In \cref{tab:train and new loss}, we present how SSC Loss influences the mIoU of training and new classes in the ADE20K-150 validation set. The results show that SSC Loss significantly improves the mIoU of new classes. This indicates that this loss could help our model learn a more consistent semantic structure of CLIP and gain stronger generalization ability for new classes.

\noindent \textbf{Additional Image Backbone.}
We show the advantage of using SAM as the additional image backbone in \cref{tab:additional vision backbone}. We only change the additional image backbone in our model, other settings are the same as our default settings.
% The results show that without an additional image backbone, the model only gets 26.3\% mIoU on ADE20K-150. 
% After we add a trainable Swin-T backbone, the mIoU is improved by 2\% . 
The results show that with a trainable Swin-T backbone, the model gets 28.5\% mIoU on ADE20K-150.
If the Swin-T backbone is frozen, the performance will be further improved by 0.5\% mIoU. When we replace the Swin-T with a Swin-B model, the mIoU is improved from 29.0\% to 29.3\%. Finally, a frozen SAM-B backbone brings a further improvement of 0.7\% mIoU compared to a frozen Swin-B. The experiment results in \cref{tab:additional vision backbone} demonstrate that the SAM-B model can help our model perform better by complementing the spacial information. The results also denote that SAM-B is better than Swin-B as an additional image backbone in the open-vocabulary semantic segmentation task.

\begin{table}[]
    \centering
    \begin{tabular}{c|c|c|c}
     \toprule
     AIB & freeze & parameters (M) & mIoU \\
     \midrule
       % None& - & 23.6 & 26.3 \\
       Swin-T \cite{liu2021swin} & $\times$ & 50.2 & 28.5 \\
       Swin-T \cite{liu2021swin}& \checkmark & 24.5 & 29.0 \\
       Swin-B \cite{liu2021swin}& \checkmark & 24.9 & 29.3 \\
       SAM-B \cite{kirillov2023segment}& \checkmark & 26.6 & \textbf{30.0 } \\
     \bottomrule
    \end{tabular}
    \caption{The ablation study on the additional image backbone (AIB). We only change the additional image backbone, the other settings are the same as our default settings.}
    \label{tab:additional vision backbone}
\end{table}

% \begin{figure*}
%   \centering
%   \begin{subfigure}{0.15\linewidth}
%     % \fbox{\rule{0pt}{2in} \rule{.9\linewidth}{0pt}}
%     \includegraphics[width=1.2\linewidth]{figures/ADE_val_00000001.jpg}
%     \caption{}
%     \label{fig:visual-a}
%   \end{subfigure}
%     \hfill
%   \begin{subfigure}{0.15\linewidth}
%     % \fbox{\rule{0pt}{2in} \rule{.9\linewidth}{0pt}}
%     \includegraphics[width=1.2\linewidth]{figures/ADE_val_00000001.jpg}
%     \caption{}
%     \label{fig:visual-b}
%   \end{subfigure}
%     \hfill
%   \begin{subfigure}{0.15\linewidth}
%     % \fbox{\rule{0pt}{2in} \rule{.9\linewidth}{0pt}}
%     \includegraphics[width=1.2\linewidth]{figures/ADE_val_00000001.jpg}
%     \caption{}
%     \label{fig:visual-c}
%   \end{subfigure}
%     \hfill
%   \begin{subfigure}{0.15\linewidth}
%     % \fbox{\rule{0pt}{2in} \rule{.9\linewidth}{0pt}}
%     \includegraphics[width=1.2\linewidth]{figures/ADE_val_00000001.jpg}
%     \caption{}
%     \label{fig:visual-d}
%   \end{subfigure}
  
%   \caption{Visualization expamples of our model on ADE20K-150.}
%   \label{fig:visualization}
% \end{figure*}

\section{Conclusion}
\label{sec:conclusion}

In this paper, to overcome the challenge that training on semantic segmentation datasets often makes the model overfit to the classes in those datasets, we propose a novel framework for open-vocabulary semantic segmentation called EBSeg, which incorporates an Adaptively Balanced Decoder (AdaB Decoder) and a Semantic Structure Consistency loss (SSC Loss). By adaptively balancing different image embeddings, AdaB Decoder can fully leverage their ability to recognize training classes and the generalization capability for new classes. The SSC Loss aligns the inter-classes affinity in the image feature space with that in the text feature space of CLIP. This loss helps our model learn a consistent semantic structure from CLIP and improves the generalization ability of our model. We also propose to fuse the image features of SAM and CLIP to complement the spatial information of CLIP image features. Our method EBSeg establishes a new state-of-the-art in the open-vocabulary semantic segmentation task.

\noindent \textbf{Acknowledgement.} This work was supported by Hubei Provincial Natural Science Foundation of China No.2022CFA055.

{
    \small
    \bibliographystyle{ieeenat_fullname}
    \bibliography{main}

\begin{thebibliography}{45}
\providecommand{\natexlab}[1]{#1}
\providecommand{\url}[1]{\texttt{#1}}
\expandafter\ifx\csname urlstyle\endcsname\relax
  \providecommand{\doi}[1]{doi: #1}\else
  \providecommand{\doi}{doi: \begingroup \urlstyle{rm}\Url}\fi

\bibitem[Bucher et~al.(2019)Bucher, Vu, Cord, and P{\'e}rez]{bucher2019zero}
Maxime Bucher, Tuan-Hung Vu, Matthieu Cord, and Patrick P{\'e}rez.
\newblock Zero-shot semantic segmentation.
\newblock \emph{Advances in Neural Information Processing Systems}, 32, 2019.

\bibitem[Caesar et~al.(2018)Caesar, Uijlings, and Ferrari]{caesar2018coco}
Holger Caesar, Jasper Uijlings, and Vittorio Ferrari.
\newblock Coco-stuff: Thing and stuff classes in context.
\newblock In \emph{Proceedings of the IEEE conference on computer vision and pattern recognition}, pages 1209--1218, 2018.

\bibitem[Chen et~al.(2018)Chen, Zhu, Papandreou, Schroff, and Adam]{chen2018encoder}
Liang-Chieh Chen, Yukun Zhu, George Papandreou, Florian Schroff, and Hartwig Adam.
\newblock Encoder-decoder with atrous separable convolution for semantic image segmentation.
\newblock In \emph{Proceedings of the European conference on computer vision (ECCV)}, pages 801--818, 2018.

\bibitem[Chen et~al.(2015)Chen, Fang, Lin, Vedantam, Gupta, Doll{\'a}r, and Zitnick]{chen2015microsoft}
Xinlei Chen, Hao Fang, Tsung-Yi Lin, Ramakrishna Vedantam, Saurabh Gupta, Piotr Doll{\'a}r, and C~Lawrence Zitnick.
\newblock Microsoft coco captions: Data collection and evaluation server.
\newblock \emph{arXiv preprint arXiv:1504.00325}, 2015.

\bibitem[Cheng et~al.(2021)Cheng, Schwing, and Kirillov]{cheng2021per}
Bowen Cheng, Alex Schwing, and Alexander Kirillov.
\newblock Per-pixel classification is not all you need for semantic segmentation.
\newblock \emph{Advances in Neural Information Processing Systems}, 34:\penalty0 17864--17875, 2021.

\bibitem[Cheng et~al.(2022)Cheng, Misra, Schwing, Kirillov, and Girdhar]{cheng2022masked}
Bowen Cheng, Ishan Misra, Alexander~G Schwing, Alexander Kirillov, and Rohit Girdhar.
\newblock Masked-attention mask transformer for universal image segmentation.
\newblock In \emph{Proceedings of the IEEE/CVF conference on computer vision and pattern recognition}, pages 1290--1299, 2022.

\bibitem[Ding et~al.(2021)Ding, Xue, Xia, and Dai]{ding2021decoupling}
Jian Ding, Nan Xue, Guisong Xia, and Dengxin Dai.
\newblock Decoupling zero-shot semantic segmentation. 2022 ieee.
\newblock In \emph{CVF Conference on Computer Vision and Pattern Recognition (CVPR)}, pages 11573--11582, 2021.

\bibitem[Ding et~al.(2023)Ding, Wang, and Tu]{ding2023open}
Zheng Ding, Jieke Wang, and Zhuowen Tu.
\newblock Open-vocabulary universal image segmentation with maskclip.
\newblock In \emph{Proceedings of the 40th International Conference on Machine Learning}, pages 8090--8102, 2023.

\bibitem[Everingham and Winn(2012)]{everingham2012pascal}
Mark Everingham and John Winn.
\newblock The pascal visual object classes challenge 2012 (voc2012) development kit.
\newblock \emph{Pattern Anal. Stat. Model. Comput. Learn., Tech. Rep}, 2007\penalty0 (1-45):\penalty0 5, 2012.

\bibitem[Ghiasi et~al.(2022)Ghiasi, Gu, Cui, and Lin]{ghiasi2022scaling}
Golnaz Ghiasi, Xiuye Gu, Yin Cui, and Tsung-Yi Lin.
\newblock Scaling open-vocabulary image segmentation with image-level labels.
\newblock In \emph{European Conference on Computer Vision}, pages 540--557. Springer, 2022.

\bibitem[Gu et~al.(2021)Gu, Lin, Kuo, and Cui]{gu2021zero}
Xiuye Gu, Tsung-Yi Lin, Weicheng Kuo, and Yin Cui.
\newblock Zero-shot detection via vision and language knowledge distillation.
\newblock \emph{arXiv preprint arXiv:2104.13921}, 2\penalty0 (3):\penalty0 4, 2021.

\bibitem[Hamilton et~al.(2021)Hamilton, Zhang, Hariharan, Snavely, and Freeman]{hamilton2021unsupervised}
Mark Hamilton, Zhoutong Zhang, Bharath Hariharan, Noah Snavely, and William~T Freeman.
\newblock Unsupervised semantic segmentation by distilling feature correspondences.
\newblock In \emph{International Conference on Learning Representations}, 2021.

\bibitem[Jia et~al.(2021)Jia, Yang, Xia, Chen, Parekh, Pham, Le, Sung, Li, and Duerig]{jia2021scaling}
Chao Jia, Yinfei Yang, Ye Xia, Yi-Ting Chen, Zarana Parekh, Hieu Pham, Quoc Le, Yun-Hsuan Sung, Zhen Li, and Tom Duerig.
\newblock Scaling up visual and vision-language representation learning with noisy text supervision.
\newblock In \emph{International conference on machine learning}, pages 4904--4916. PMLR, 2021.

\bibitem[Kirillov et~al.(2019{\natexlab{a}})Kirillov, Girshick, He, and Doll{\'a}r]{kirillov2019panoptic}
Alexander Kirillov, Ross Girshick, Kaiming He, and Piotr Doll{\'a}r.
\newblock Panoptic feature pyramid networks.
\newblock In \emph{Proceedings of the IEEE/CVF conference on computer vision and pattern recognition}, pages 6399--6408, 2019{\natexlab{a}}.

\bibitem[Kirillov et~al.(2019{\natexlab{b}})Kirillov, He, Girshick, Rother, and Doll{\'a}r]{kirillov2019panoptic1}
Alexander Kirillov, Kaiming He, Ross Girshick, Carsten Rother, and Piotr Doll{\'a}r.
\newblock Panoptic segmentation.
\newblock In \emph{Proceedings of the IEEE/CVF conference on computer vision and pattern recognition}, pages 9404--9413, 2019{\natexlab{b}}.

\bibitem[Kirillov et~al.(2023)Kirillov, Mintun, Ravi, Mao, Rolland, Gustafson, Xiao, Whitehead, Berg, Lo, et~al.]{kirillov2023segment}
Alexander Kirillov, Eric Mintun, Nikhila Ravi, Hanzi Mao, Chloe Rolland, Laura Gustafson, Tete Xiao, Spencer Whitehead, Alexander~C Berg, Wan-Yen Lo, et~al.
\newblock Segment anything.
\newblock \emph{arXiv preprint arXiv:2304.02643}, 2023.

\bibitem[Lester et~al.(2021)Lester, Al-Rfou, and Constant]{lester2021power}
Brian Lester, Rami Al-Rfou, and Noah Constant.
\newblock The power of scale for parameter-efficient prompt tuning.
\newblock In \emph{Proceedings of the 2021 Conference on Empirical Methods in Natural Language Processing}, pages 3045--3059, 2021.

\bibitem[Li et~al.(2022)Li, Weinberger, Belongie, Koltun, and Ranftl]{li2022languagedriven}
Boyi Li, Kilian~Q. Weinberger, Serge Belongie, Vladlen Koltun, and René Ranftl.
\newblock Language-driven semantic segmentation, 2022.

\bibitem[Li and Liang(2021)]{li2021prefix}
Xiang~Lisa Li and Percy Liang.
\newblock Prefix-tuning: Optimizing continuous prompts for generation.
\newblock \emph{arXiv preprint arXiv:2101.00190}, 2021.

\bibitem[Li et~al.(2023)Li, Wang, Duan, and Li]{li2023clip}
Yi Li, Hualiang Wang, Yiqun Duan, and Xiaomeng Li.
\newblock Clip surgery for better explainability with enhancement in open-vocabulary tasks, 2023.

\bibitem[Liang et~al.(2023)Liang, Wu, Dai, Li, Zhao, Zhang, Zhang, Vajda, and Marculescu]{liang2023open}
Feng Liang, Bichen Wu, Xiaoliang Dai, Kunpeng Li, Yinan Zhao, Hang Zhang, Peizhao Zhang, Peter Vajda, and Diana Marculescu.
\newblock Open-vocabulary semantic segmentation with mask-adapted clip.
\newblock In \emph{Proceedings of the IEEE/CVF Conference on Computer Vision and Pattern Recognition}, pages 7061--7070, 2023.

\bibitem[Liang-Chieh et~al.(2015)Liang-Chieh, Papandreou, Kokkinos, Murphy, and Yuille]{liang2015semantic}
Chen Liang-Chieh, George Papandreou, Iasonas Kokkinos, Kevin Murphy, and Alan Yuille.
\newblock Semantic image segmentation with deep convolutional nets and fully connected crfs.
\newblock In \emph{International Conference on Learning Representations}, 2015.

\bibitem[Liu et~al.(2021)Liu, Lin, Cao, Hu, Wei, Zhang, Lin, and Guo]{liu2021swin}
Ze Liu, Yutong Lin, Yue Cao, Han Hu, Yixuan Wei, Zheng Zhang, Stephen Lin, and Baining Guo.
\newblock Swin transformer: Hierarchical vision transformer using shifted windows.
\newblock In \emph{Proceedings of the IEEE/CVF international conference on computer vision}, pages 10012--10022, 2021.

\bibitem[Long et~al.(2015)Long, Shelhamer, and Darrell]{long2015fully}
Jonathan Long, Evan Shelhamer, and Trevor Darrell.
\newblock Fully convolutional networks for semantic segmentation.
\newblock In \emph{Proceedings of the IEEE conference on computer vision and pattern recognition}, pages 3431--3440, 2015.

\bibitem[Loshchilov and Hutter(2017)]{loshchilov2017decoupled}
Ilya Loshchilov and Frank Hutter.
\newblock Decoupled weight decay regularization.
\newblock \emph{arXiv preprint arXiv:1711.05101}, 2017.

\bibitem[Mottaghi et~al.(2014)Mottaghi, Chen, Liu, Cho, Lee, Fidler, Urtasun, and Yuille]{mottaghi2014role}
Roozbeh Mottaghi, Xianjie Chen, Xiaobai Liu, Nam-Gyu Cho, Seong-Whan Lee, Sanja Fidler, Raquel Urtasun, and Alan Yuille.
\newblock The role of context for object detection and semantic segmentation in the wild.
\newblock In \emph{Proceedings of the IEEE conference on computer vision and pattern recognition}, pages 891--898, 2014.

\bibitem[Noh et~al.(2015)Noh, Hong, and Han]{noh2015learning}
Hyeonwoo Noh, Seunghoon Hong, and Bohyung Han.
\newblock Learning deconvolution network for semantic segmentation.
\newblock In \emph{Proceedings of the IEEE international conference on computer vision}, pages 1520--1528, 2015.

\bibitem[Paszke et~al.(2019)Paszke, Gross, Massa, Lerer, Bradbury, Chanan, Killeen, Lin, Gimelshein, Antiga, et~al.]{paszke2019pytorch}
Adam Paszke, Sam Gross, Francisco Massa, Adam Lerer, James Bradbury, Gregory Chanan, Trevor Killeen, Zeming Lin, Natalia Gimelshein, Luca Antiga, et~al.
\newblock Pytorch: An imperative style, high-performance deep learning library.
\newblock \emph{Advances in neural information processing systems}, 32, 2019.

\bibitem[Pont-Tuset et~al.(2020)Pont-Tuset, Uijlings, Changpinyo, Soricut, and Ferrari]{pont2020connecting}
Jordi Pont-Tuset, Jasper Uijlings, Soravit Changpinyo, Radu Soricut, and Vittorio Ferrari.
\newblock Connecting vision and language with localized narratives.
\newblock In \emph{Computer Vision--ECCV 2020: 16th European Conference, Glasgow, UK, August 23--28, 2020, Proceedings, Part V 16}, pages 647--664. Springer, 2020.

\bibitem[Radford et~al.(2021)Radford, Kim, Hallacy, Ramesh, Goh, Agarwal, Sastry, Askell, Mishkin, Clark, et~al.]{radford2021learning}
Alec Radford, Jong~Wook Kim, Chris Hallacy, Aditya Ramesh, Gabriel Goh, Sandhini Agarwal, Girish Sastry, Amanda Askell, Pamela Mishkin, Jack Clark, et~al.
\newblock Learning transferable visual models from natural language supervision.
\newblock In \emph{International conference on machine learning}, pages 8748--8763. PMLR, 2021.

\bibitem[Rao et~al.(2022)Rao, Zhao, Chen, Tang, Zhu, Huang, Zhou, and Lu]{rao2022denseclip}
Yongming Rao, Wenliang Zhao, Guangyi Chen, Yansong Tang, Zheng Zhu, Guan Huang, Jie Zhou, and Jiwen Lu.
\newblock Denseclip: Language-guided dense prediction with context-aware prompting.
\newblock In \emph{Proceedings of the IEEE/CVF Conference on Computer Vision and Pattern Recognition}, pages 18082--18091, 2022.

\bibitem[Rombach et~al.(2022)Rombach, Blattmann, Lorenz, Esser, and Ommer]{rombach2022high}
Robin Rombach, Andreas Blattmann, Dominik Lorenz, Patrick Esser, and Bj{\"o}rn Ommer.
\newblock High-resolution image synthesis with latent diffusion models.
\newblock In \emph{Proceedings of the IEEE/CVF conference on computer vision and pattern recognition}, pages 10684--10695, 2022.

\bibitem[Tan and Le(2019)]{tan2019efficientnet}
Mingxing Tan and Quoc Le.
\newblock Efficientnet: Rethinking model scaling for convolutional neural networks.
\newblock In \emph{International conference on machine learning}, pages 6105--6114. PMLR, 2019.

\bibitem[Wu et~al.(2019)Wu, Kirillov, Massa, Lo, and Girshick]{wu2019detectron2}
Yuxin Wu, Alexander Kirillov, Francisco Massa, Wan-Yen Lo, and Ross Girshick.
\newblock Detectron2.
\newblock \url{https://github.com/facebookresearch/detectron2}, 2019.

\bibitem[Xian et~al.(2019)Xian, Choudhury, He, Schiele, and Akata]{xian2019semantic}
Yongqin Xian, Subhabrata Choudhury, Yang He, Bernt Schiele, and Zeynep Akata.
\newblock Semantic projection network for zero-and few-label semantic segmentation.
\newblock In \emph{Proceedings of the IEEE/CVF Conference on Computer Vision and Pattern Recognition}, pages 8256--8265, 2019.

\bibitem[Xiao et~al.(2018)Xiao, Liu, Zhou, Jiang, and Sun]{xiao2018unified}
Tete Xiao, Yingcheng Liu, Bolei Zhou, Yuning Jiang, and Jian Sun.
\newblock Unified perceptual parsing for scene understanding.
\newblock In \emph{Proceedings of the European conference on computer vision (ECCV)}, pages 418--434, 2018.

\bibitem[Xie et~al.(2021)Xie, Wang, Yu, Anandkumar, Alvarez, and Luo]{xie2021segformer}
Enze Xie, Wenhai Wang, Zhiding Yu, Anima Anandkumar, Jose~M Alvarez, and Ping Luo.
\newblock Segformer: Simple and efficient design for semantic segmentation with transformers.
\newblock \emph{Advances in Neural Information Processing Systems}, 34:\penalty0 12077--12090, 2021.

\bibitem[Xu et~al.(2023{\natexlab{a}})Xu, Liu, Vahdat, Byeon, Wang, and De~Mello]{xu2023open}
Jiarui Xu, Sifei Liu, Arash Vahdat, Wonmin Byeon, Xiaolong Wang, and Shalini De~Mello.
\newblock Open-vocabulary panoptic segmentation with text-to-image diffusion models.
\newblock In \emph{Proceedings of the IEEE/CVF Conference on Computer Vision and Pattern Recognition}, pages 2955--2966, 2023{\natexlab{a}}.

\bibitem[Xu et~al.(2022)Xu, Zhang, Wei, Lin, Cao, Hu, and Bai]{xu2022simple}
Mengde Xu, Zheng Zhang, Fangyun Wei, Yutong Lin, Yue Cao, Han Hu, and Xiang Bai.
\newblock A simple baseline for open-vocabulary semantic segmentation with pre-trained vision-language model.
\newblock In \emph{European Conference on Computer Vision}, pages 736--753. Springer, 2022.

\bibitem[Xu et~al.(2023{\natexlab{b}})Xu, Zhang, Wei, Hu, and Bai]{xu2023side}
Mengde Xu, Zheng Zhang, Fangyun Wei, Han Hu, and Xiang Bai.
\newblock Side adapter network for open-vocabulary semantic segmentation.
\newblock In \emph{Proceedings of the IEEE/CVF Conference on Computer Vision and Pattern Recognition}, pages 2945--2954, 2023{\natexlab{b}}.

\bibitem[Zhao et~al.(2017)Zhao, Puig, Zhou, Fidler, and Torralba]{zhao2017open}
Hang Zhao, Xavier Puig, Bolei Zhou, Sanja Fidler, and Antonio Torralba.
\newblock Open vocabulary scene parsing.
\newblock In \emph{Proceedings of the IEEE International Conference on Computer Vision}, pages 2002--2010, 2017.

\bibitem[Zhong et~al.(2021)Zhong, Friedman, and Chen]{zhong2021factual}
Zexuan Zhong, Dan Friedman, and Danqi Chen.
\newblock Factual probing is [mask]: Learning vs. learning to recall.
\newblock \emph{arXiv preprint arXiv:2104.05240}, 2021.

\bibitem[Zhou et~al.(2017)Zhou, Zhao, Puig, Fidler, Barriuso, and Torralba]{zhou2017scene}
Bolei Zhou, Hang Zhao, Xavier Puig, Sanja Fidler, Adela Barriuso, and Antonio Torralba.
\newblock Scene parsing through ade20k dataset.
\newblock In \emph{Proceedings of the IEEE conference on computer vision and pattern recognition}, pages 633--641, 2017.

\bibitem[Zhou et~al.(2022{\natexlab{a}})Zhou, Yang, Loy, and Liu]{zhou2022conditional}
Kaiyang Zhou, Jingkang Yang, Chen~Change Loy, and Ziwei Liu.
\newblock Conditional prompt learning for vision-language models.
\newblock In \emph{Proceedings of the IEEE/CVF Conference on Computer Vision and Pattern Recognition}, pages 16816--16825, 2022{\natexlab{a}}.

\bibitem[Zhou et~al.(2022{\natexlab{b}})Zhou, Yang, Loy, and Liu]{zhou2022learning}
Kaiyang Zhou, Jingkang Yang, Chen~Change Loy, and Ziwei Liu.
\newblock Learning to prompt for vision-language models.
\newblock \emph{International Journal of Computer Vision}, 130\penalty0 (9):\penalty0 2337--2348, 2022{\natexlab{b}}.

\end{thebibliography}
}

% \clearpage
% \setcounter{page}{1}
% \maketitlesupplementary

\setcounter{section}{0}
\renewcommand\thesection{\Alph{section}}

\begin{figure*}
  \centering
  \begin{subfigure}{0.15\linewidth}
    % \fbox{\rule{0pt}{2in} \rule{.9\linewidth}{0pt}}
    \includegraphics[width=1.2\linewidth,height=1\linewidth]{figures/visual/compare/input/1.jpg}
    \includegraphics[width=1.2\linewidth]{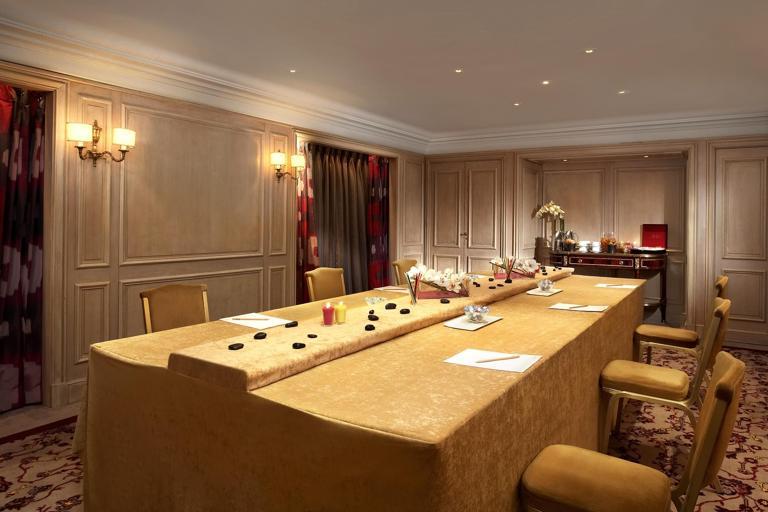}
     \includegraphics[width=1.2\linewidth]{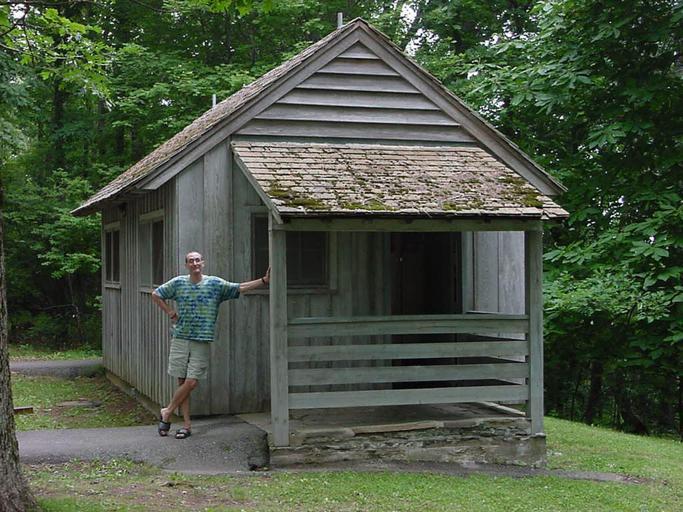}
     \includegraphics[width=1.2\linewidth]{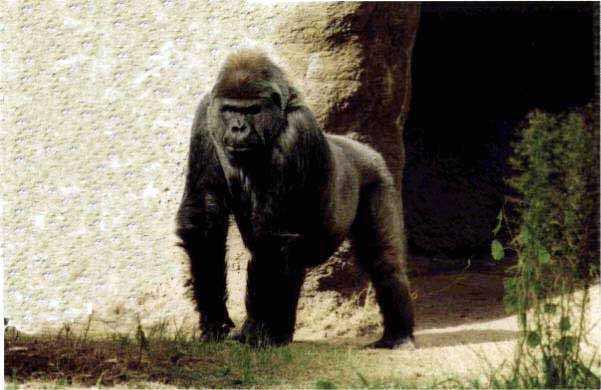}
    \caption{Input}
    \label{fig:visual-a}
  \end{subfigure}
  \hspace{0.5cm}
  \begin{subfigure}{0.15\linewidth}
    % \fbox{\rule{0pt}{2in} \rule{.9\linewidth}{0pt}}
    \includegraphics[width=1.2\linewidth,height=1\linewidth]{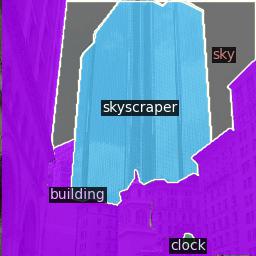}
    \includegraphics[width=1.2\linewidth]{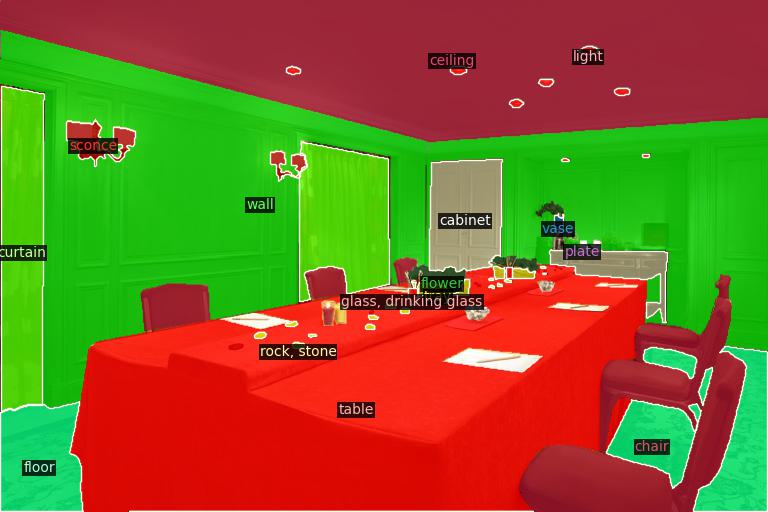}
     \includegraphics[width=1.2\linewidth]{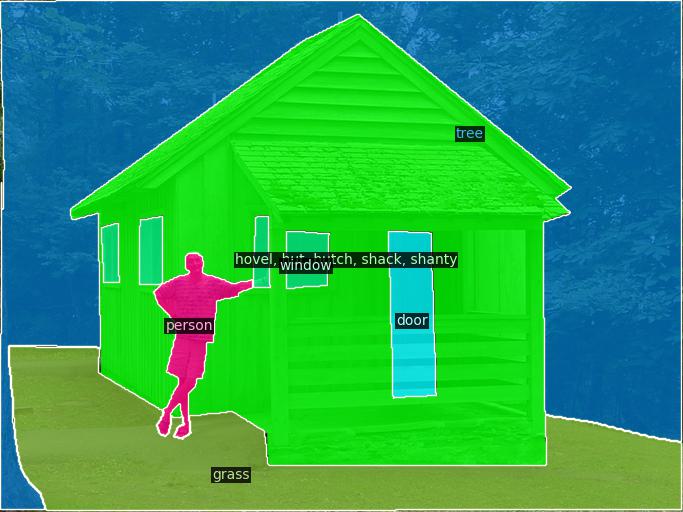}
     \includegraphics[width=1.2\linewidth]{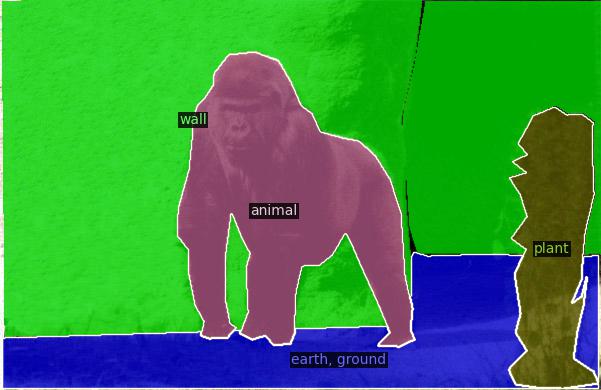}
    \caption{GT}
    \label{fig:visual-b}
  \end{subfigure}
    \hspace{0.5cm}
  \begin{subfigure}{0.15\linewidth}
    % \fbox{\rule{0pt}{2in} \rule{.9\linewidth}{0pt}}
  \includegraphics[width=1.2\linewidth,height=1\linewidth]{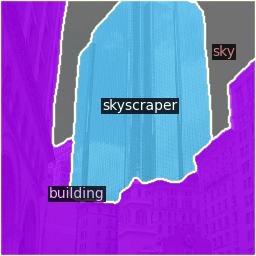}
    \includegraphics[width=1.2\linewidth]{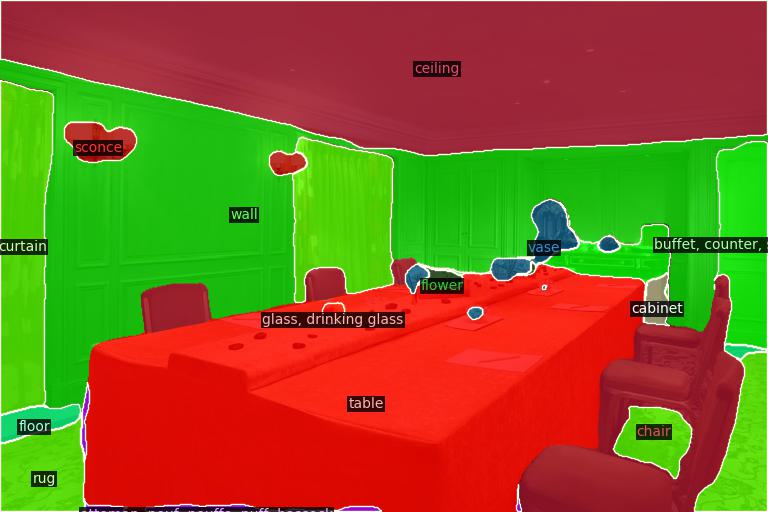}
     \includegraphics[width=1.2\linewidth]{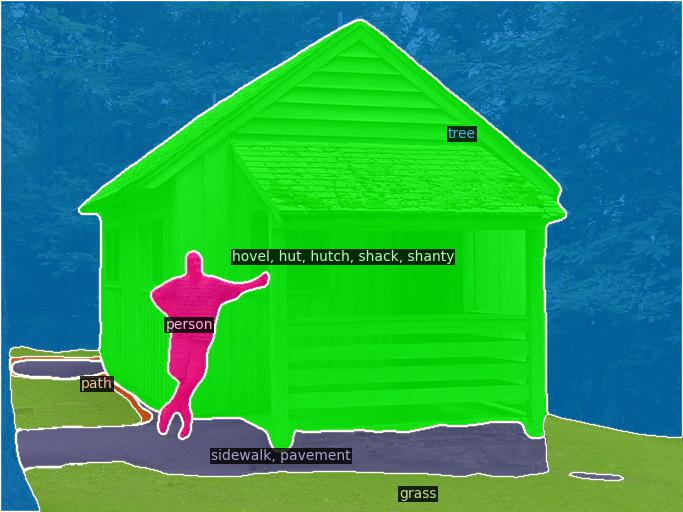}
     \includegraphics[width=1.2\linewidth]{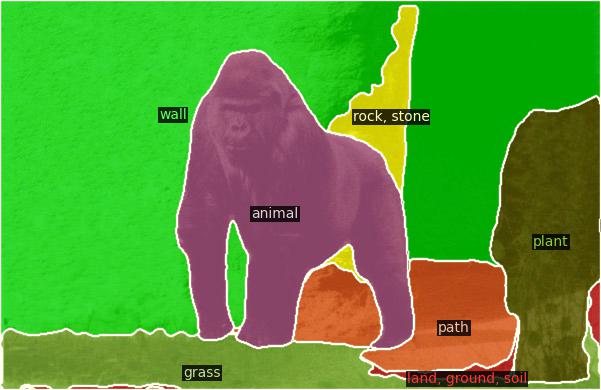}
    \caption{\textbf{EBSeg(ours)}}
    \label{fig:visual-c}
  \end{subfigure}
  \hspace{0.5cm}
  \begin{subfigure}{0.15\linewidth}
    % \fbox{\rule{0pt}{2in} \rule{.9\linewidth}{0pt}}
    \includegraphics[width=1.2\linewidth,height=1\linewidth]{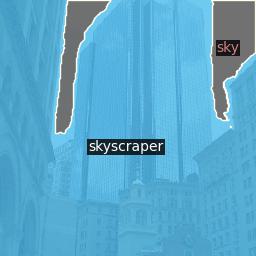}
    \includegraphics[width=1.2\linewidth]{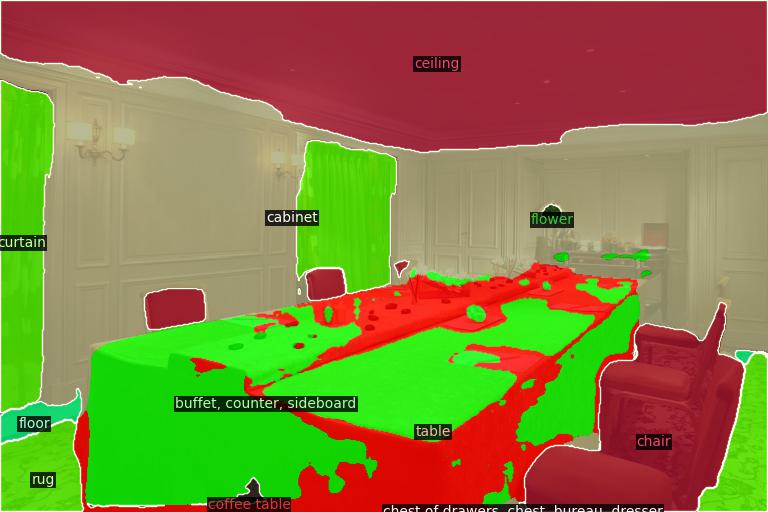}
     \includegraphics[width=1.2\linewidth]{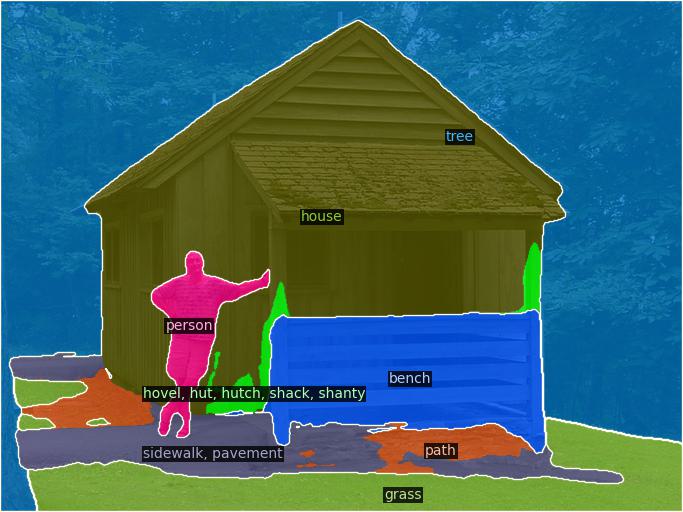}
     \includegraphics[width=1.2\linewidth]{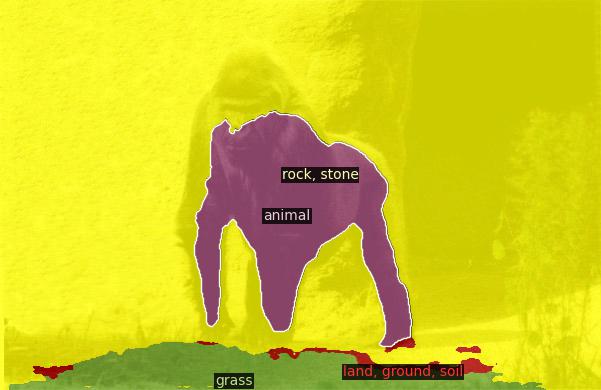}
    \caption{OVSeg \cite{liang2023open}}
    \label{fig:visual-e}
  \end{subfigure}
  \hspace{0.5cm}
  \begin{subfigure}{0.15\linewidth}
    % \fbox{\rule{0pt}{2in} \rule{.9\linewidth}{0pt}}
    \includegraphics[width=1.2\linewidth,height=1\linewidth]{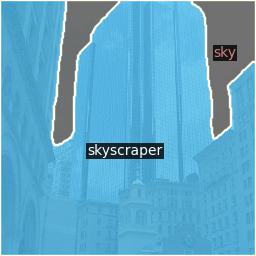}
    \includegraphics[width=1.2\linewidth]{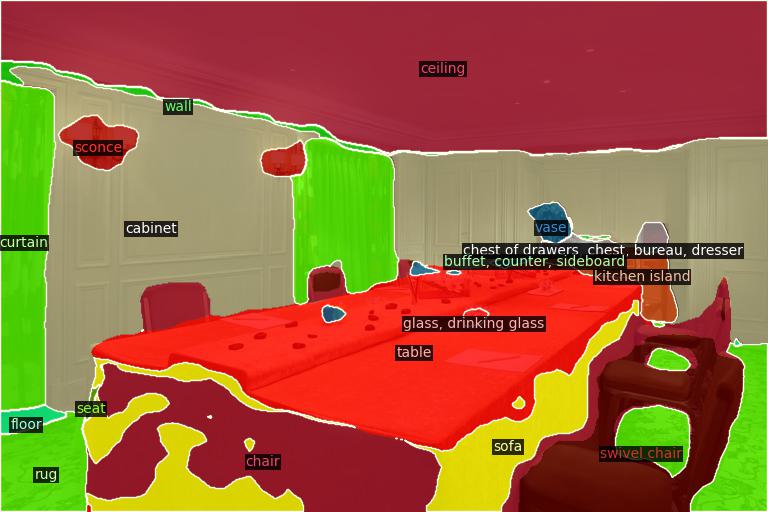}
     \includegraphics[width=1.2\linewidth]{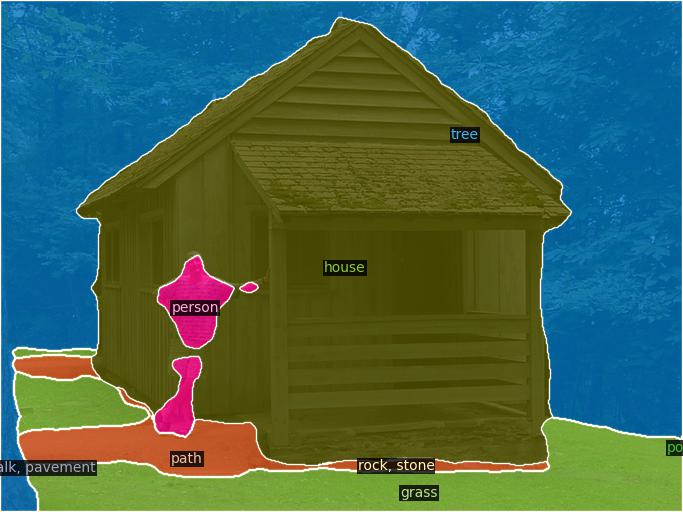}
     \includegraphics[width=1.2\linewidth]{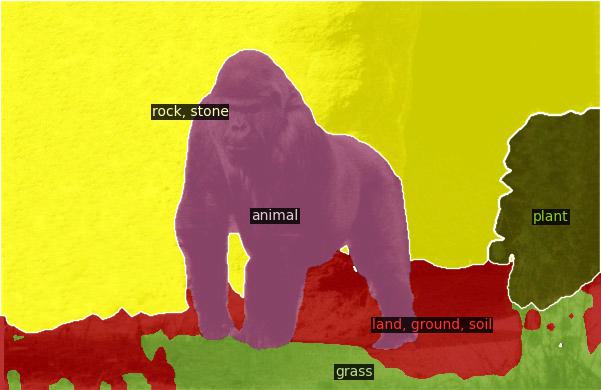}
    \caption{SAN \cite{xu2023side}}
    \label{fig:visual-f}
  \end{subfigure}
  \caption{Qualitative results on the ADE20K-150 \cite{zhou2017scene} validation set. We compare our approach with two other methods OVSeg \cite{xu2023side} and SAN \cite{liang2023open}. Thanks to our AdaB Decoder and SSC Loss, our model shows a stronger generalization ability for new classes that do not exist in the training dataset COCO-Stuff \cite{caesar2018coco}, such as \textit{hovel} in the third row and \textit{animal} in the last row. Moreover, with the help of our AdaB Decoder, our model is able to better recognize training classes that exist in the training set, such as \textit{building} in the first row and \textit{table}, \textit{wall} in the second row.}
  \label{fig:visualization compare}
\end{figure*}

\section{Additional Experiments}
\label{more experi}
% In this section, we conducted some additional experiments.
All results are obtained with CLIP \cite{radford2021learning} ViT-B/16 based model and on the ADE20K-150 \cite{zhou2017scene} validation set.

\subsection{Embedding Balancing Strategy and Weight}
\label{weight}
In the main paper, we adaptively balance the three image embeddings $\mathbf{A}$, $\mathbf{B}$ and $\mathbf{D} \in \mathbb{R}^{N \times C}$ by computing their geometric mean. This subsection focuses on studying the effects of different averaging strategies (arithmetic mean and geometric mean) and weights ($\alpha$, $\beta$ and $\gamma$) on the mIoU. 

When we take the geometric mean strategy, since there are negative values in the image embeddings $\mathbf{A}$, $\mathbf{B}$, and $\mathbf{D}$, we cannot directly take their geometric mean. Actually, our geometric mean implementation follows ODISE \cite{xu2023open}, getting different geometric mean of prediction logits for different classes. We first calculate the prediction logits of all the embeddings with all classes:
\begin{equation}
  \mathbf{P}_{A} = \mathbf{A} \times \mathbf{TE}_{test}^T,
  \label{eq:logit_A}
\end{equation}
\begin{equation}
  \mathbf{P}_{B} = \mathbf{B} \times \mathbf{TE}_{test}^T,
  \label{eq:logit_B}
\end{equation}
\begin{equation}
  \mathbf{P}_{D} = \mathbf{D} \times \mathbf{TE}_{test}^T,
  \label{eq:logit_D}
\end{equation}
where $\mathbf{P}_{A}$, $\mathbf{P}_{B}$ and $\mathbf{P}_{D}\in \mathbb{R}^{N \times K}$.
Then, we combine the prediction logits of those three kinds of embeddings with geometric mean for training and new classes:
\begin{equation}
  \mathbf{P}_{train} = \mathbf{P}_{A}^{\alpha} \odot \mathbf{P}_{B}^{\beta} \odot \mathbf{P}_{D}^{1-\alpha-\beta},
  \label{eq:logit_traing}
\end{equation}
\begin{equation}
  \mathbf{P}_{new} = \mathbf{P}_{A}^{\gamma} \odot \mathbf{P}_{B}^{\beta} \odot \mathbf{P}_{D}^{1-\gamma-\beta},
  \label{eq:logit_new}
\end{equation}
where $\mathbf{P}_{train}$ and $\mathbf{P}_{new} \in \mathbb{R}^{N \times K}$. After that, we remove the columns (the second dimension) corresponding to the new categories in $\mathbf{P}_{train}$, as well as the columns corresponding to the training classes in $\mathbf{P}_{new}$, getting the same $\mathbf{P}_{train} \in \mathbb{R}^{N \times f}$ and $\mathbf{P}_{new} \in \mathbb{R}^{N \times (K-f)}$ in the main paper.
%因为image embeddings \mathbf{A}, \mathbf{B} and \mathbf{D}中有负值，所以无法直接对它们取几何平均值，所以我们先使用这三种embeddings各自计算和所有类别的prediction logits，然后再将不同的logit做组合。具体过程如下，首先我们获取三种prediction logits：，然后将取这三种prediction logits的几何平均值，得到不同类别的prediction logits：我们将...中的新类别对应的列以及...中的training class对应的列去掉，得到和正文中一样的...和...
% , the balanced embeddings for training and new classes are:
% \begin{equation}
%   \mathbf{E}_{train} = \alpha * \mathbf{A} + \beta * \mathbf{B} + (1-\alpha-\beta) * \mathbf{D},
%   \label{eq:a_mean_embed_1}
% \end{equation}
% \begin{equation}
%   \mathbf{E}_{new} = \gamma * \mathbf{A} + \beta * \mathbf{B} + (1-\gamma-\beta) * \mathbf{D},
%   \label{eq:a_mean_embed_2}
% \end{equation}
% where $\mathbf{E}_{train}$, $\mathbf{E}_{new} \in \mathbb{R}^{N \times C}$.

In \cref{tab:balancing weight}, we conduct experiments with different image embedding balancing strategies and weights during inference. At first, we set $\alpha=0.4$, $\beta=0.6$, $\gamma=0$. Then we adjust the the three weights and choose a best set of weights for our model.
We adjust the weights in the order of $\alpha$, $\gamma$, $\beta$.
The experiments show that geometric mean is slightly better for the image embedding balancing, so we use geometric mean in our default setting.

\begin{table}[]
    \centering
    \begin{tabular}{c|c|c|c|c|c}
     \toprule
     adjusted weight& $\alpha$ & $\beta$ & $\gamma$ & mIoU$_{g}$ & mIoU$_{a}$\\
     \midrule
   \multirow{4}{*}{\centering $\alpha$}   &0.4 & 0.6 & 0.0 & 29.3&29.1\\
      &0.3 & 0.6 & 0.0 & 29.7&29.4\\
     &0.2 & 0.6 & 0.0 & \textbf{29.8}&29.6\\
     &0.1 & 0.6 & 0.0 & 29.2&28.8\\
      \midrule
     \multirow{3}{*}{\centering $\gamma$} &0.2 & 0.6 & 0.0 & \textbf{29.8}&29.6\\
        &0.2 & 0.6 & 0.1 & 29.6&29.1\\
       &0.2 & 0.6 & 0.2 & 29.2&28.8\\
      \midrule
      \multirow{5}{*}{\centering $\beta$} &0.2 & 0.6 & 0.0 & 29.8&29.6\\
      &0.2 & 0.5 & 0.0 & 29.2&28.7\\
       &0.2 & 0.4 & 0.0 & 28.6&28.2\\
       &0.2 & 0.7 & 0.0 & \textbf{30.0}&29.7\\
       &0.2 & 0.8 & 0.0 & 29.6&29.5\\
     \bottomrule
    \end{tabular}
    \caption{Ablation study on the image embedding balancing strategy and weight on ADE20K-150 \cite{zhou2017scene} validation set. mIoU$_{g}$ denotes the mIoU results using geometric mean for embedding balancing. mIoU$_{a}$ is the mIoU results with arithmetic mean.}
    \label{tab:balancing weight}
\end{table}

\begin{figure*}
  \centering
  \begin{subfigure}{0.15\linewidth}
    % \fbox{\rule{0pt}{2in} \rule{.9\linewidth}{0pt}}
    \includegraphics[width=1.2\linewidth]{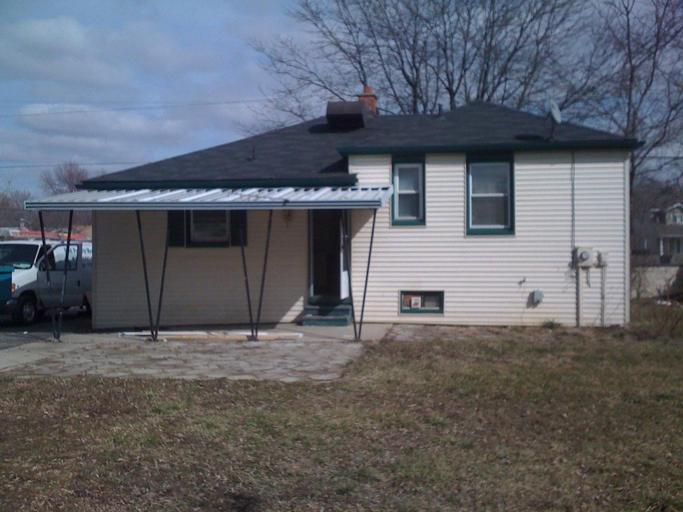}
    \includegraphics[width=1.2\linewidth]{figures/visual/ade-150-847/input/2.jpg}
    \includegraphics[width=1.2\linewidth]{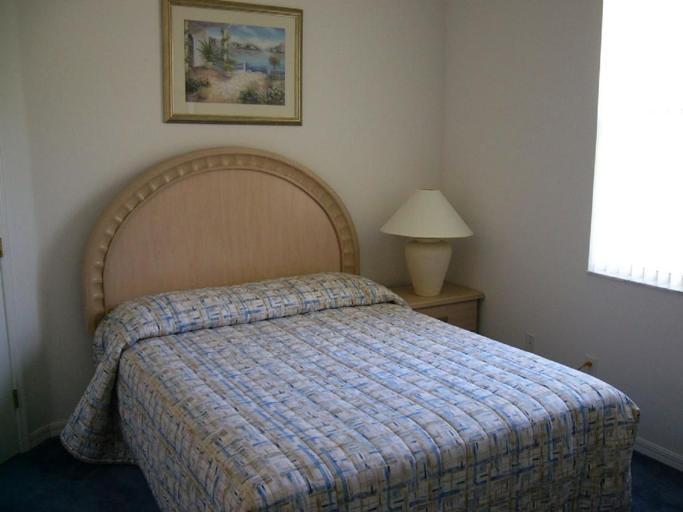}
    \includegraphics[width=1.2\linewidth]{figures/visual/ade-150-847/input/4.jpg}
    \caption{Input}
    \label{fig:visual-a1}
  \end{subfigure}
  \hspace{0.5cm}
  \begin{subfigure}{0.15\linewidth}
    % \fbox{\rule{0pt}{2in} \rule{.9\linewidth}{0pt}}
    \includegraphics[width=1.2\linewidth]{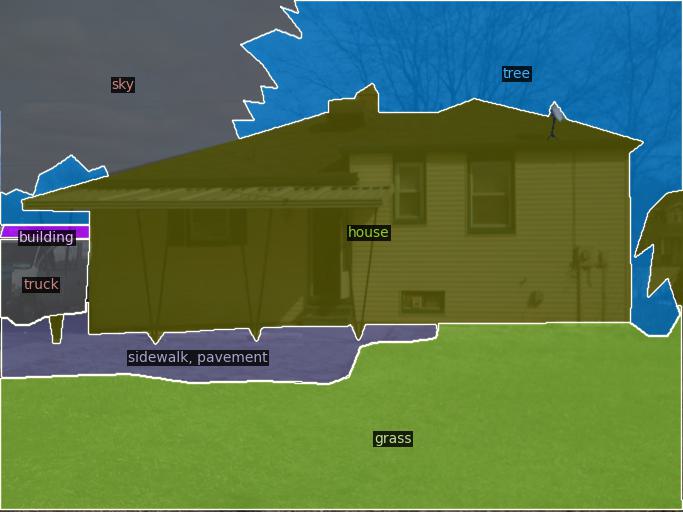}
     \includegraphics[width=1.2\linewidth]{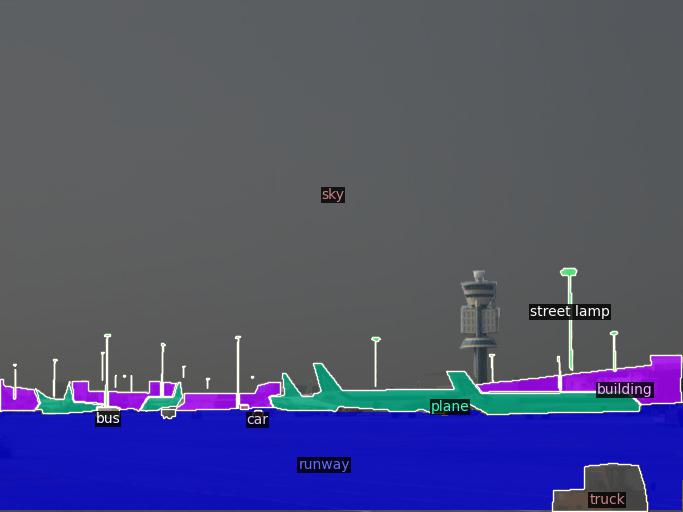}
      \includegraphics[width=1.2\linewidth]{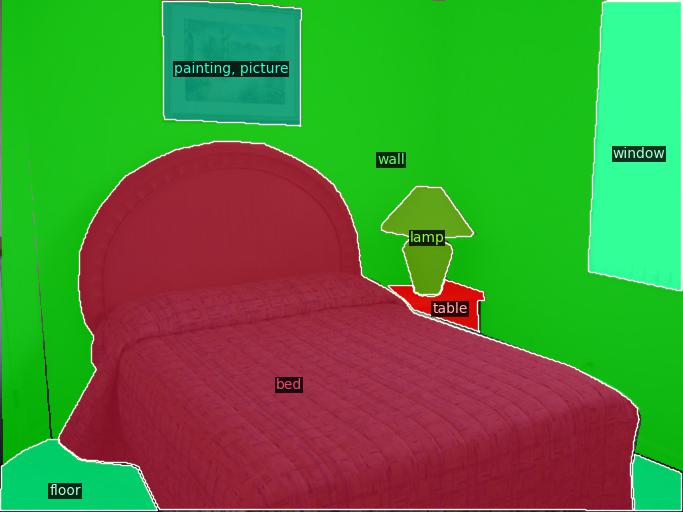}
       \includegraphics[width=1.2\linewidth]{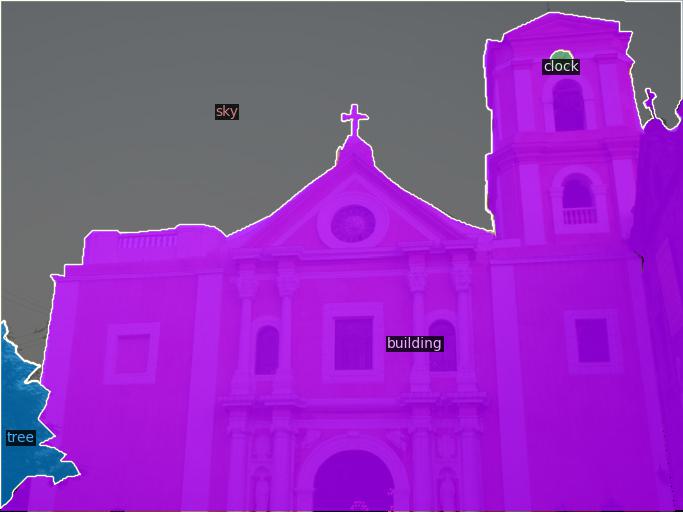}
    \caption{GT (150)}
    \label{fig:visual-b1}
  \end{subfigure}
    \hspace{0.5cm}
  \begin{subfigure}{0.15\linewidth}
    % \fbox{\rule{0pt}{2in} \rule{.9\linewidth}{0pt}}
  \includegraphics[width=1.2\linewidth]{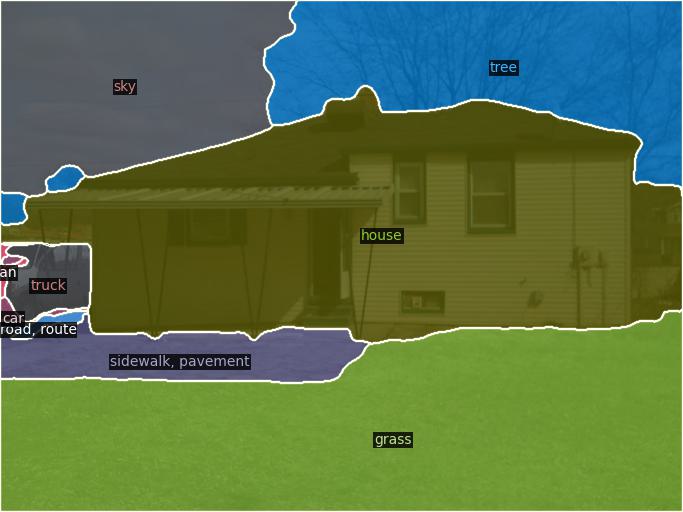}
    \includegraphics[width=1.2\linewidth]{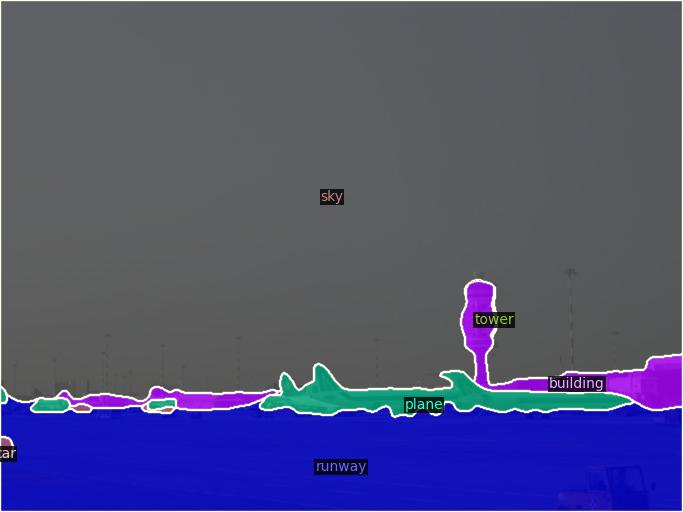}
    \includegraphics[width=1.2\linewidth]{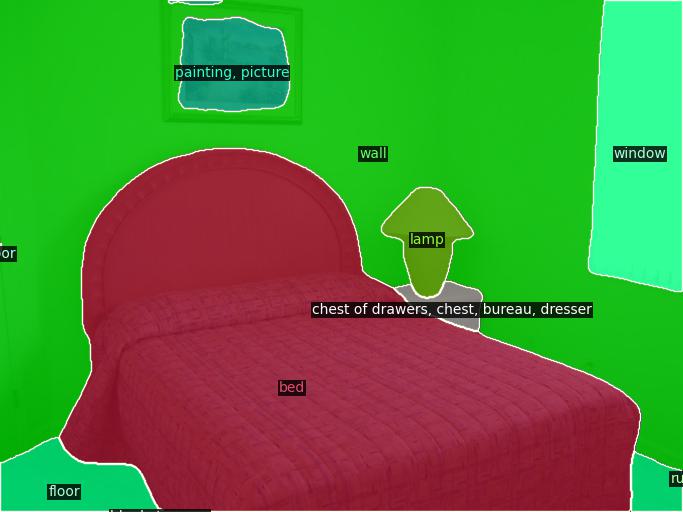}
    \includegraphics[width=1.2\linewidth]{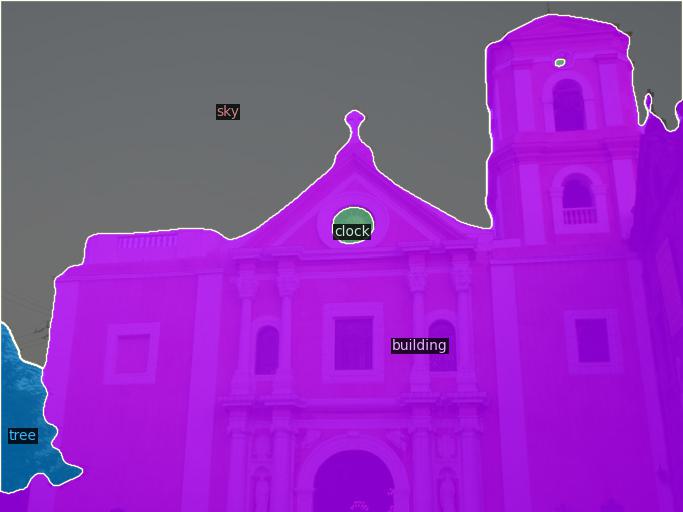}
    \caption{prediction (150)}
    \label{fig:visual-c1}
  \end{subfigure}
  \hspace{0.5cm}
  \begin{subfigure}{0.15\linewidth}
    % \fbox{\rule{0pt}{2in} \rule{.9\linewidth}{0pt}}
    \includegraphics[width=1.2\linewidth]{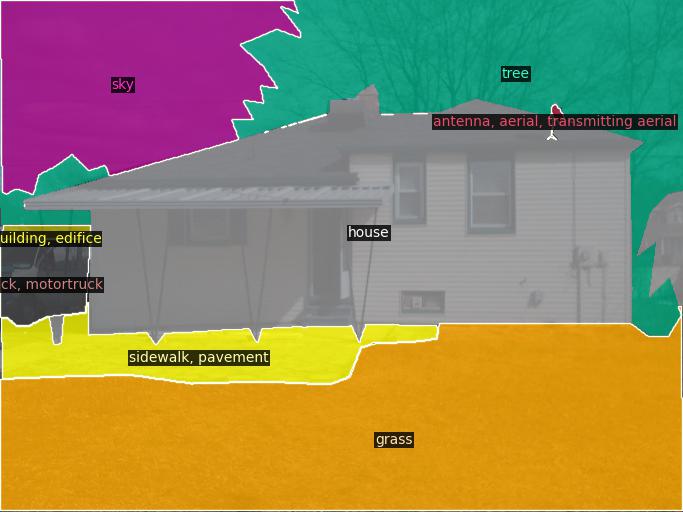}
    \includegraphics[width=1.2\linewidth]{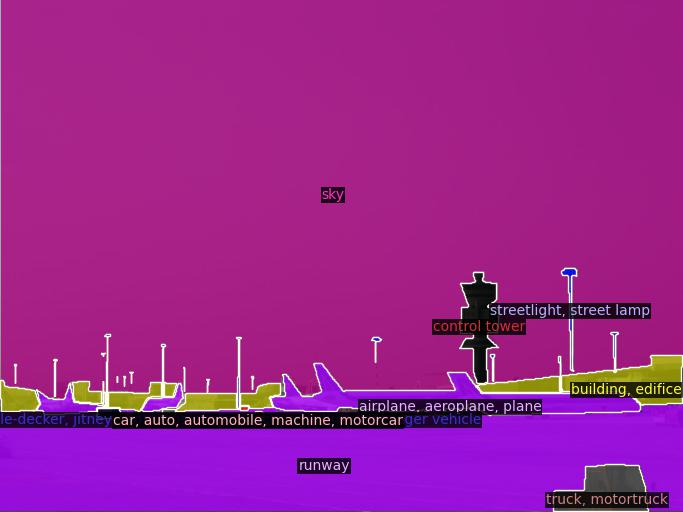}
   \includegraphics[width=1.2\linewidth]{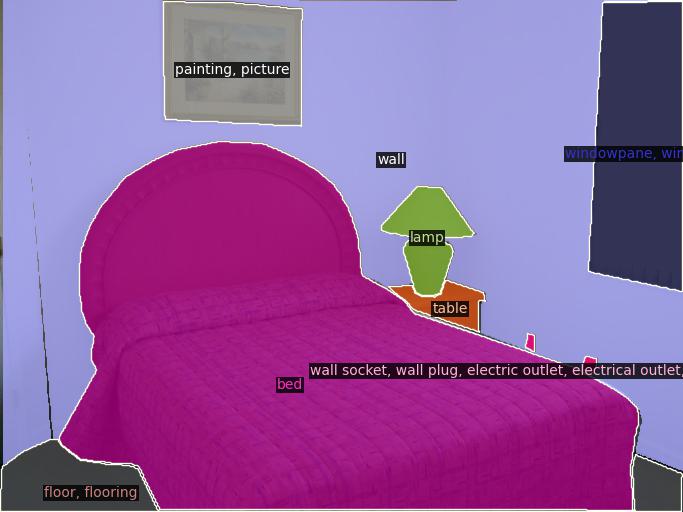}
   \includegraphics[width=1.2\linewidth]{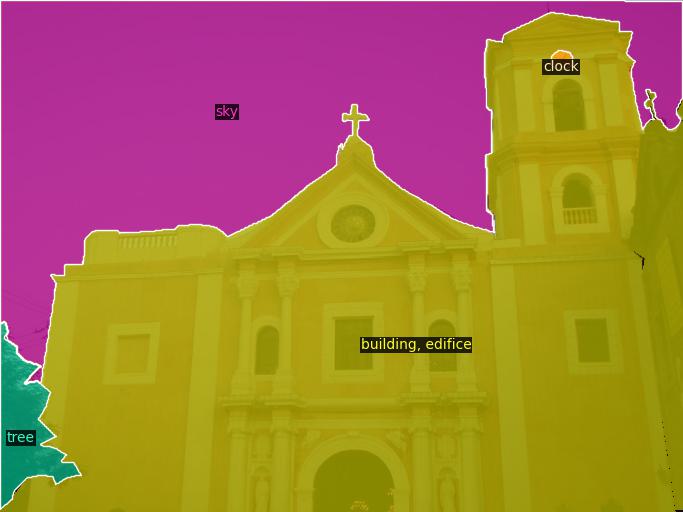}
    \caption{GT (847)}
    \label{fig:visual-e1}
  \end{subfigure}
  \hspace{0.5cm}
  \begin{subfigure}{0.15\linewidth}
    % \fbox{\rule{0pt}{2in} \rule{.9\linewidth}{0pt}}
    \includegraphics[width=1.2\linewidth]{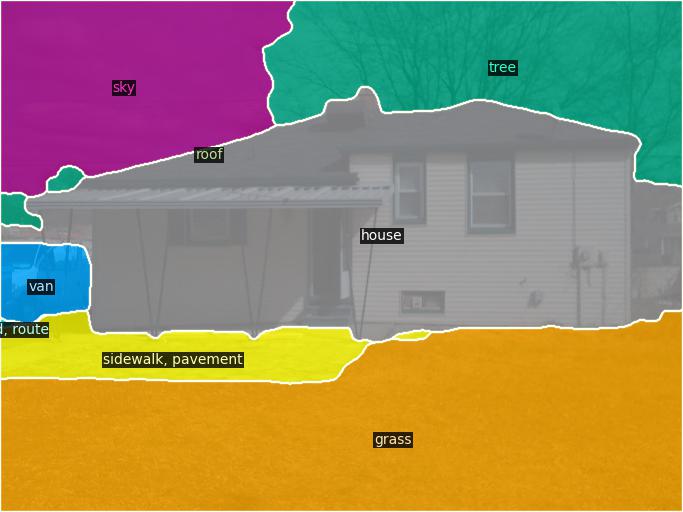}
    \includegraphics[width=1.2\linewidth]{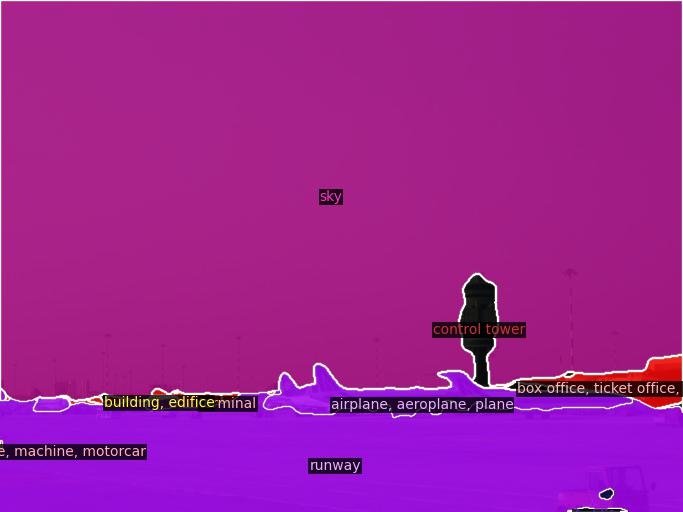}
        \includegraphics[width=1.2\linewidth]{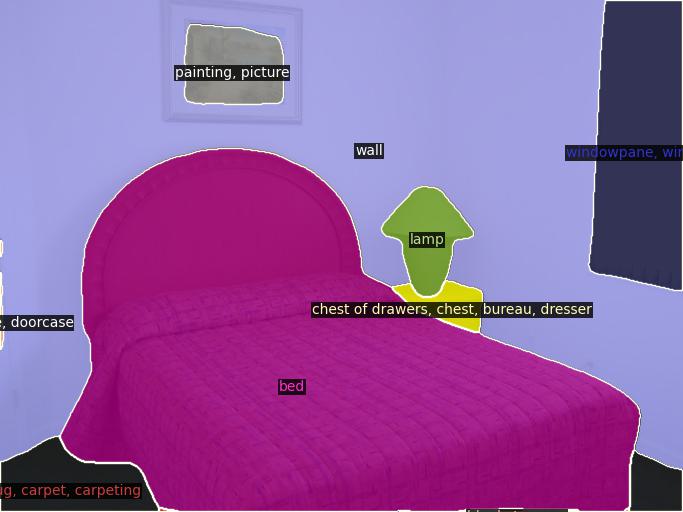}
            \includegraphics[width=1.2\linewidth]{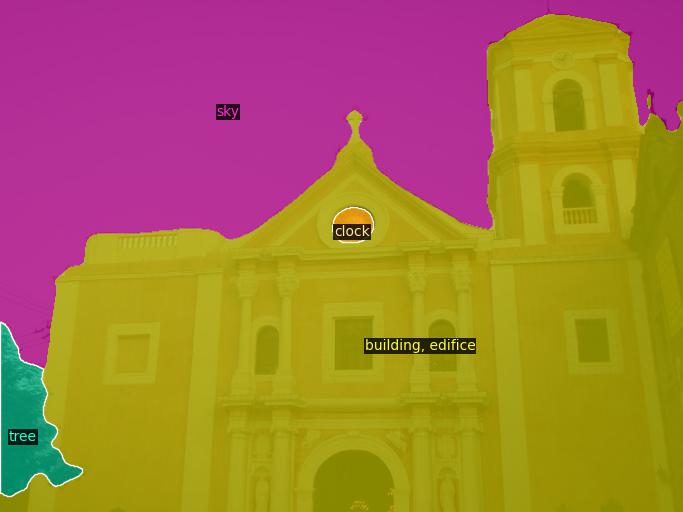}
    \caption{prediction (847)}
    \label{fig:visual-f1}
  \end{subfigure}
%换成更大的词汇量就比更多的类别更好了
  \caption{Qualitative results on the ADE20K-150/847 \cite{zhou2017scene} validation set. ADE20K-847 has a broader vocabulary than ADE20K-150.}
  \label{fig:visualization ade}
\end{figure*}

\section{More Implement Details}
\label{more implement detail}

\subsection{Text Prompt}
\label{prompt engineering}

\begin{table}[]
    \centering
    \begin{tabular}{c|c}
     \toprule
     prompt strategy & mIoU\\
     \midrule
       single template & 29.1 \\
       prompt engineering &  \textbf{30.0} \\
       prompt tuning & 25.3 \\
     \bottomrule
    \end{tabular}
    \caption{Ablation study on text prompt strategies. \textit{single prompt} denotes using one prompt template: "a photo of a \{\}.". \textit{prompt engineering} means using the 14 prompt templates in ViLD \cite{gu2021zero}. \textit{prompt tuning} indicates using trainable prompt embeddings introduced in CoOp \cite{zhou2022conditional}. We use the CLIP ViT-B/16 based model and evaluate our model on the ADE20K-150 validation set.}
    \label{tab:prompt engineering}
\end{table}

% \textbf{Prompt engineering.}
Following previous works \cite{xu2023side, xu2023open}, we use the ViLD \cite{gu2021zero} prompt templates to extract text embeddings. There are 14 prompt templates in ViLD, such as  "a photo of a \{\}." and "There is a \{\} in the scene". We use all these 14 templates to extract text embeddings and average the text embeddings of different templates for each class to get the final text embeddings for training and inference.
We show the results of different prompt strategies in \cref{tab:prompt engineering}. When applying prompt tuning, since the trainable prompt embeddings easily overfit to the training classes, the mIoU drops significantly. Prompt engineering with ViLD prompts helps our model the most.

\begin{table}[]
    \centering
    \begin{tabular}{c|c|c|c}
     \toprule
        CLIP model & method & params (M) & GFLOPs \\
     \midrule
       \multirow{2}{*}{\centering ViT-B/16} & baseline & 23.5 & 339.9 \\
        & EBSeg & \textbf{26.6} &\textbf{312.3} \\
     \midrule
       \multirow{2}{*}{\centering ViT-L/14} & baseline & 24.9 &  1132.4\\
        & EBSeg & \textbf{28.0} & \textbf{622.1} \\
     \bottomrule
    \end{tabular}
    \caption{Model size of EBSeg. \textit{baseline} denotes that we do not use the additional image backbone and do not downsample the input for CLIP image encoder. The GFLOPs are measured with input images of $640^2$ resolution.}
    \label{tab:model size}
\end{table}

\subsection{The Semantic Segmentation Loss}
\label{all losses}
During training, following Mask2former \cite{cheng2022masked}, we apply binary cross-entropy loss $L_{bce}$ and dice loss $L_{dice}$ to supervise the mask ($\mathbf{M}$) generation process, and apply cross-entropy loss $L_{cls}$ to supervise the mask classification process. Note that we apply $L_{cls}$ to both mask classification results of fully supervised image embeddings $\mathbf{A}$ and mask attention image embeddings $\mathbf{B}$. So our semantic segmentation loss $L_{sem\_seg}$ is:
\begin{equation}
  L_{sem\_seg} = \lambda_1  L_{bce}+ \lambda_2 L_{dice}+ \lambda_3 L_{cls}.
  \label{eq:seg loss}
\end{equation}
Following the default setting in Mask2former \cite{cheng2022masked}, we set $\lambda_1=5$, $\lambda_2=5$ and $\lambda_3=2$.

\begin{figure*}[!t]
  \centering
  \begin{tabular}{ccccc}
    \rotatebox{90}{Input Image} &
      \includegraphics[height=0.14\textwidth]{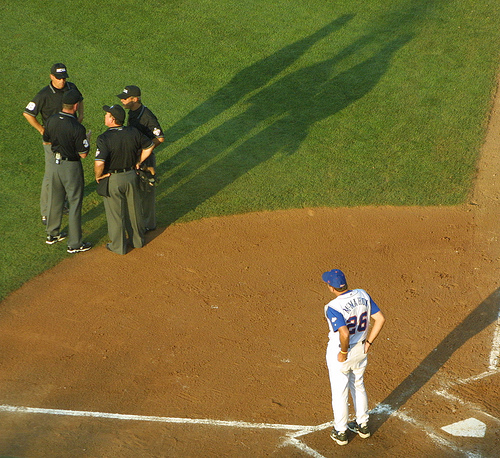} & \includegraphics[height=0.14\textwidth]{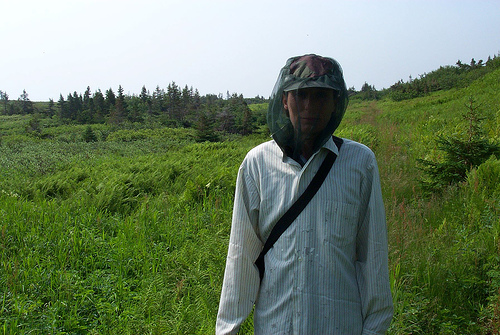} &
    \includegraphics[height=0.14\textwidth]{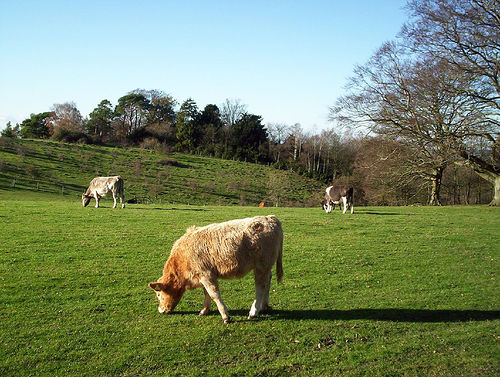} & \includegraphics[height=0.14\textwidth]{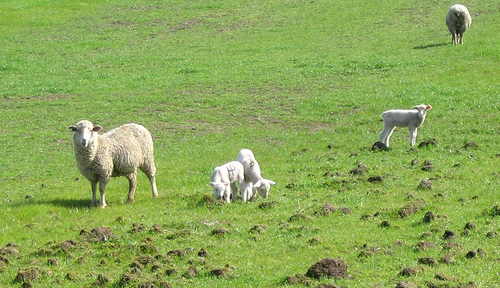} \\
  \rotatebox{90}{Ground Truth} &
  \includegraphics[height=0.14\textwidth]{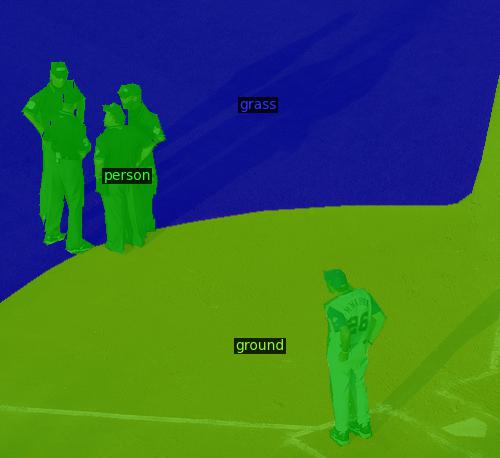} & \includegraphics[height=0.14\textwidth]{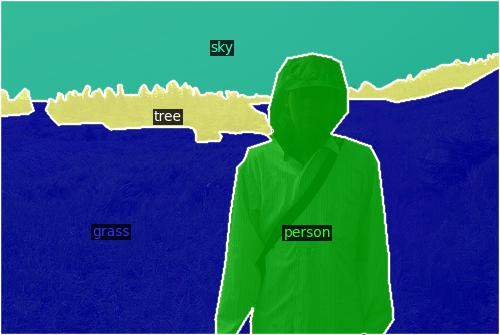} &
    \includegraphics[height=0.14\textwidth]{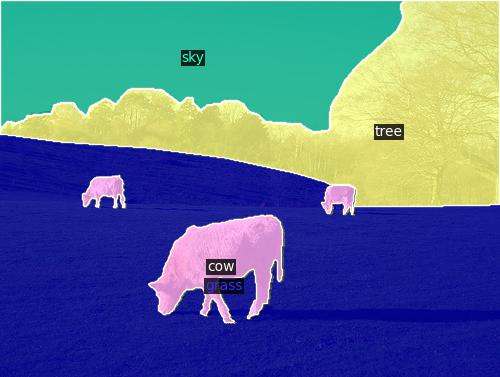} & \includegraphics[height=0.14\textwidth]{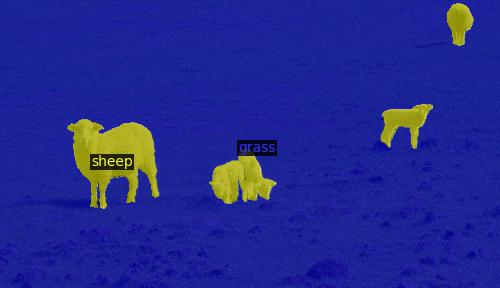} \\
\rotatebox{90}{Prediction} &
  \includegraphics[height=0.14\textwidth]{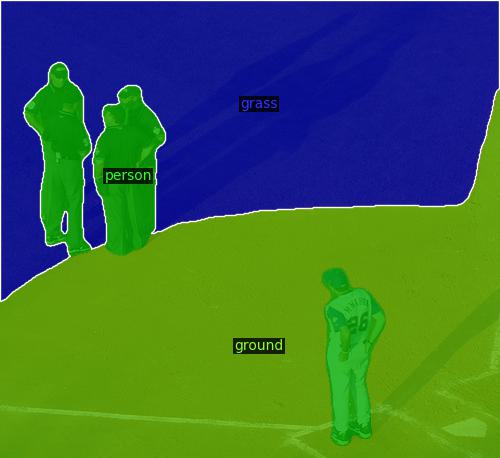} & \includegraphics[height=0.14\textwidth]{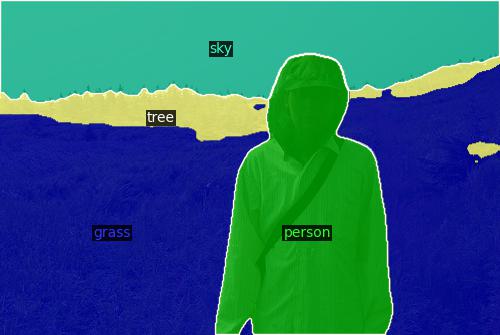}  &
    \includegraphics[height=0.14\textwidth]{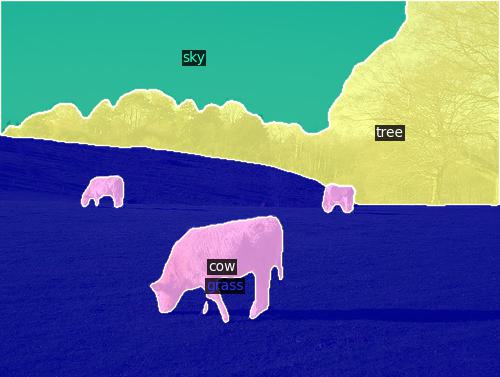} & \includegraphics[height=0.14\textwidth]{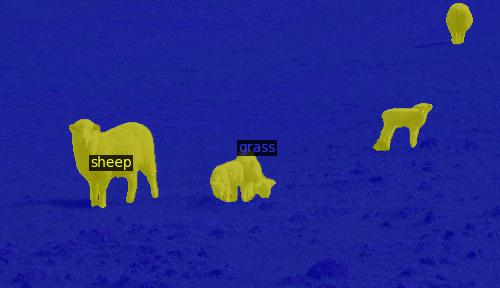}  \\
  \end{tabular}
  \caption{Qualitative results on the PC-59 \cite{mottaghi2014role} validation set.}
  \label{fig:visualization-pc59}
\end{figure*}

\begin{figure*}[!t]
  \centering
  \begin{tabular}{ccccc}
    \rotatebox{90}{Input Image} &
      \includegraphics[height=0.14\textwidth]{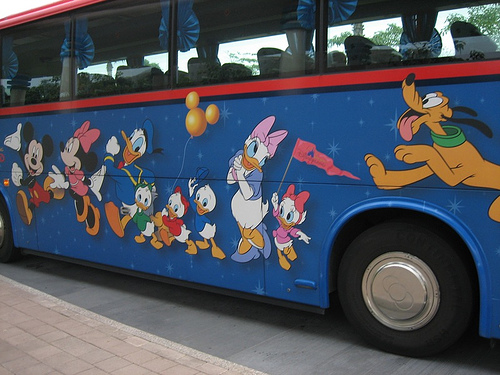} & \includegraphics[height=0.14\textwidth]{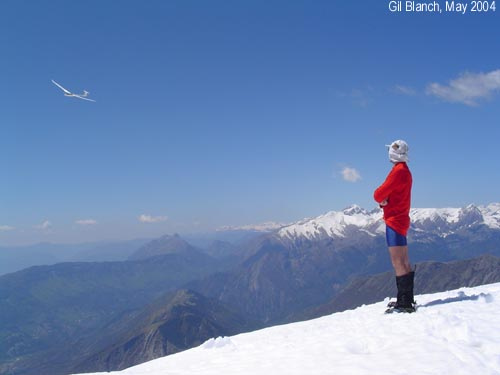} &
    \includegraphics[height=0.14\textwidth]{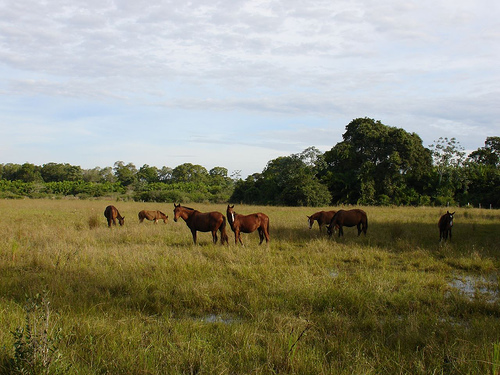} & \includegraphics[height=0.14\textwidth]{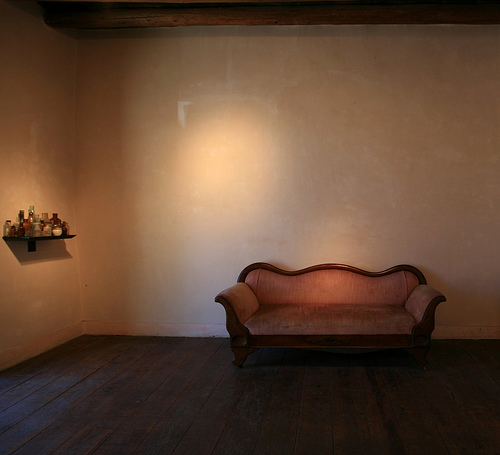}\\
  \rotatebox{90}{Ground Truth} &
  \includegraphics[height=0.14\textwidth]{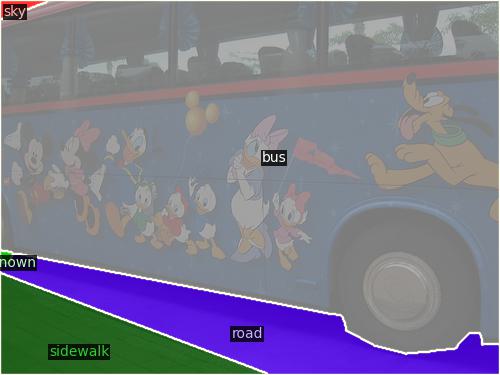} & \includegraphics[height=0.14\textwidth]{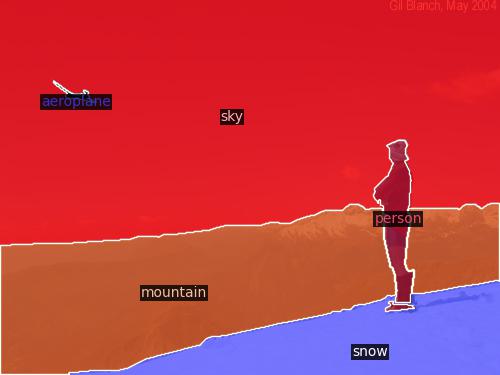} &
    \includegraphics[height=0.14\textwidth]{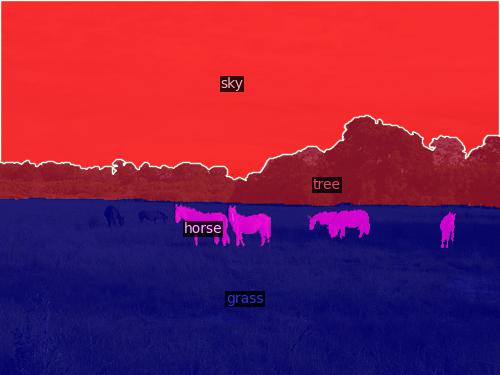}& \includegraphics[height=0.14\textwidth]{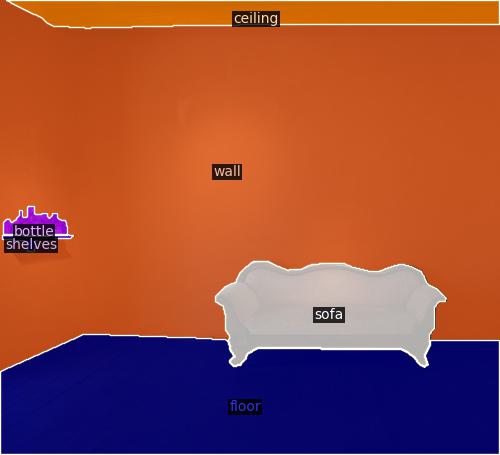} \\
\rotatebox{90}{Prediction} &
  \includegraphics[height=0.14\textwidth]{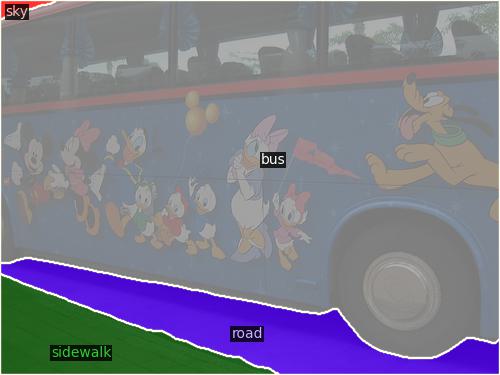} & \includegraphics[height=0.14\textwidth]{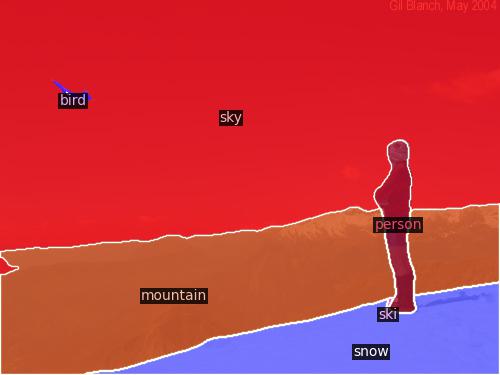} &
    \includegraphics[height=0.14\textwidth]{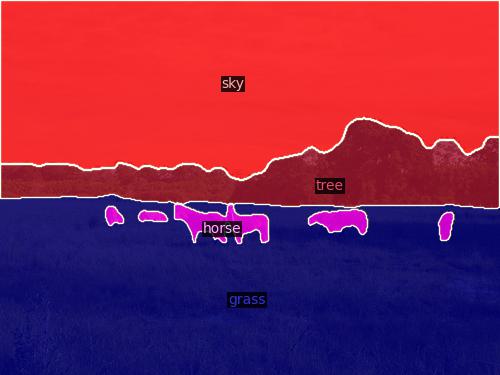}& \includegraphics[height=0.14\textwidth]{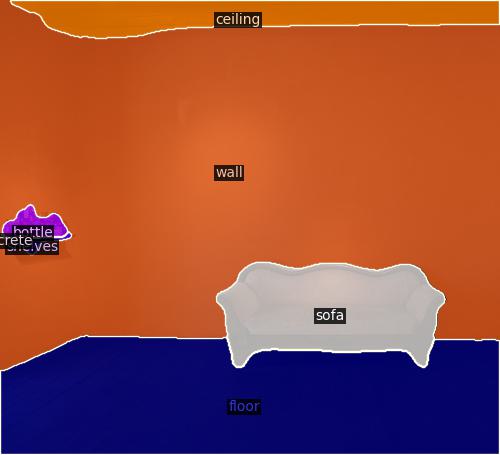} \\
  \end{tabular}
  \caption{Qualitative results on the PC-459 \cite{mottaghi2014role} validation set.}
  \label{fig:visualization-pc459}
\end{figure*}

\subsection{Model Size}
\label{model size}
We list the number of parameters and GFLOPs of our models in \cref{tab:model size}. During training, we freeze all parameters of CLIP and SAM \cite{kirillov2023segment} image encoders except for their positional embeddings. Since we downsample the input image for CLIP image encoder and use a SAM-B image encoder for all our models (both CLIP ViT-B and ViT-L based models), our models have fewer GFLOPs than a baseline model that do not use an additional image backbone. Please note that during training and inference, we only use the CLIP text encoder once to extract text embeddings in the first iteration. So when measuring GFLOPs, we do not include the  CLIP text encoder.

\section{More Qualitative Results}
\label{more visual}
In this section, we provide more qualitative results of our model to demonstrate the effectiveness of our proposed approach.
We first compare our model with other methods \cite{xu2023side, liang2023open} in \cref{fig:visualization compare} on the ADE20K-150 validation set.
Then we present more qualitative results of our model on ADE20K-150/847 in \cref{fig:visualization ade}, PC-59 \cite{mottaghi2014role} in \cref{fig:visualization-pc59} and PC-459 \cite{mottaghi2014role} in \cref{fig:visualization-pc459}. All the visualization results are output by CLIP ViT-B/16 based models.

% {
%     \small
%     \bibliographystyle{ieeenat_fullname}
%     \bibliography{main}
% }

\end{document}